\documentclass[runningheads]{llncs}

\usepackage[mobile]{eccv}

\usepackage{eccvabbrv}

\usepackage{graphicx}
\usepackage{booktabs}
\usepackage{amsmath}
\usepackage{caption}
\usepackage{float}
\usepackage{multirow}
\usepackage{multicol}
\usepackage[linesnumbered,ruled,vlined]{algorithm2e}

\usepackage[accsupp]{axessibility}  % Improves PDF readability for those with disabilities.

\newcommand{\Fmat}[0]{{{\boldsymbol F}}}

\newcommand{\Imat}{{\boldsymbol I}}

\newcommand{\Rmat}[0]{{{\boldsymbol R}}}
\newcommand{\Smat}[0]{{{\boldsymbol S}}}

\newcommand{\Ymat}[0]{{{\boldsymbol Y}}}

\usepackage{hyperref}

\usepackage{orcidlink}

\begin{document}

\title{PolarAPP: Beyond Polarization Demosaicking for Polarimetric Applications} 

\titlerunning{PolarAPP}

\author{Yidong Luo\inst{1,2}\thanks{Equal contribution. \quad $^\dagger$ Corresponding authors.}\orcidlink{0000-0002-9665-6471} \and
Chenggong Li\inst{3}\protect\footnotemark[1]\orcidlink{0009-0003-9955-642X} \and Yunfeng Song\inst{1,2}\orcidlink{0009-0001-9758-3016} \and Ping Wang\inst{2}\orcidlink{0009-0001-2746-5102} \and Boxin Shi\inst{4}\orcidlink{0000-0001-6749-0364} \and Junchao Zhang\inst{3}$^\dagger$\orcidlink{0000-0003-2243-0012} \and Xin Yuan\inst{2}$^\dagger$\orcidlink{0000-0002-8311-7524}}

\authorrunning{Y. Luo, C. Li et al.}

\institute{Zhejiang University, Hangzhou, China \and
School of Engineering, Westlake University, Hangzhou, China \\
\email{\{luoyidong, xyuan\}@westlake.edu.cn}
\and School of Automation, Central South University, Changsha, China \\
\email{\{244603040, junchaozhang\}@csu.edu.cn}
\and School of Computer Science, Peking University, Beijing, China
}

\maketitle

\begin{abstract}
  Polarimetric imaging enables advanced vision applications such as normal estimation and de-reflection by capturing unique surface-material interactions. However, existing applications (alternatively called downstream tasks) rely on datasets constructed by na\"ively regrouping raw measurements from division-of-focal-plane sensors—where pixels of the same polarization angle are extracted and aligned into sparse images without proper demosaicking. This reconstruction strategy results in suboptimal, incomplete targets that limit downstream performance. Moreover, current demosaicking methods are task-agnostic, optimizing only for photometric fidelity rather than utility in downstream tasks. Towards this end, we propose PolarAPP, the first framework to jointly optimize demosaicking and its downstream tasks. PolarAPP introduces a feature alignment mechanism that semantically aligns the representations of demosaicking and downstream networks via meta-learning, guiding the reconstruction to be task-aware. It further employs an equivalent imaging constraint for demosaicking training, enabling direct regression to physically meaningful outputs without relying on rearranged data. Finally, a task-refinement stage fine-tunes the task network using the stable demosaicking front-end to further enhance accuracy. Extensive experimental results demonstrate that PolarAPP outperforms existing methods in both demosaicking quality and downstream performance. Code is available \href{https://github.com/roydon-luo/PolarAPP}{here}. 
\end{abstract}

\section{Introduction}
\label{sec:intro}

Polarimetric imaging captures physical properties beyond RGB, such as surface normal and subsurface scattering, enabling advanced vision tasks in target detection~\cite{luo2025cpifuse}, early cancer diagnosis~\cite{tuniyazi2024snapshot} and other high-level tasks. However, off-the-shelf polarization sensors (e.g., Sony IMX250MZR) output mosaic images, where each pixel records only one polarization angle, necessitating demosaicking to reconstruct linear polarization images for downstream applications.

{\begin{figure}[t]
\centering
    \includegraphics[width=0.99\textwidth]{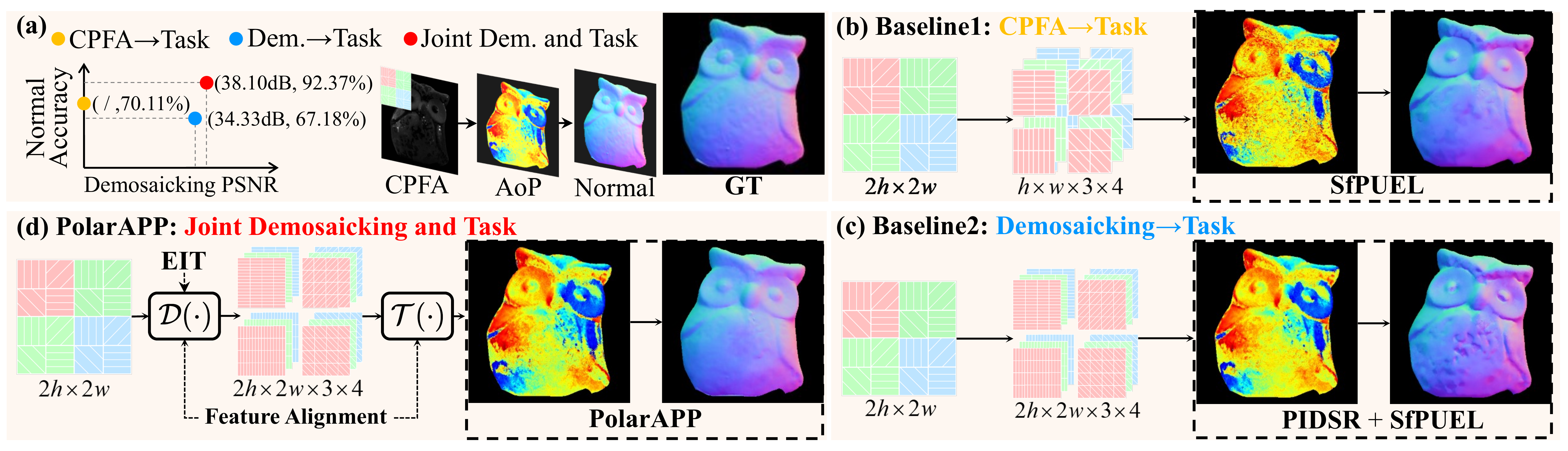}
    \captionsetup{type=figure,skip=4pt}
    \caption{(a) Comparison between two baselines and our {\bf joint PolarAPP}, which directly performs polarization-based tasks from a single CPFA raw image. (b) The current SOTA pipeline, SfPUEL~\cite{lyu2024sfpuel}, suffers from artifacts in the reconstructed normal map due to noisy AoP estimation. (c) A na\"ive combination of the SOTA demosaicker PIDSR~\cite{zhou2025pidsr} with SfPUEL still fails to improve normal estimation. (d) In contrast, our PolarAPP yields sharper AoP and a more accurate normal map through customized joint learning with equivalent imaging transformation (EIT) and feature alignment.}
\label{fig1}
\end{figure}}

Despite the growing interest in polarization vision, existing algorithm designs do not take division-of-focal-plane (DoFP) imaging into consideration. Existing methods for Shape from Polarization (SfP)~\cite{ba2020deep, lei2022shape, lyu2024sfpuel}, De-reflection from Polarization (DfP)~\cite{lyu2019reflection, lei2020polarized, yao2025polarfree} and other polarization-based tasks~\cite{zhu2024podb} are trained on datasets constructed by naively rearranging DoFP measurements into four polarization channels. Concretely, a CPFA (color polarization filter array) sensor produces a single raw mosaic (e.g., $2048\times2448$) that interleaves an RGGB Bayer pattern with a $2\times2$ DoFP micro-polarizer pattern. Prior pipelines typically ($i$) regroup pixels within each $2\times2$ DoFP block into four polarization views \{$0^\circ$, $45^\circ$, $90^\circ$, $135^\circ$\}, yielding a half-resolution tensor of size $1024\times1224\times4$, and then ($ii$) apply Bayer demosaicking each view to obtain color polarization images of size $1024\times1224\times3\times4$. While this simplification provides convenient supervision, it bypasses polarization-aware demosaicking, leaving interpolation artifacts and polarization-angle misalignment uncorrected; moreover, the $2\times2$ regrouping inevitably reduces spatial resolution. As shown in Fig.~\ref{fig1}(b), noisy AoP computed from the regrouped measurements can propagate to downstream predictions (e.g., inaccurate normals), motivating an explicit demosaicking stage that reconstructs physically consistent polarization cues for task learning.

Meanwhile, current deep learning-based demosaickers~\cite{sun2021color, nguyen2022two, guo2024attention, li2025demosaicking, zhou2025pidsr} are primarily designed for image reconstruction, optimizing for pixel-level or perceptual fidelity metrics such as LPIPS~\cite{zhang2018unreasonable}. Although these objectives enhance visual quality, they overlook the semantic requirements of downstream tasks. Consequently, the demosaicked images may appear visually correct yet remain not task-ready. Moreover, when demosaicker and task models are trained separately, the demosaicker receives no task-level feedback and thus cannot adapt its reconstruction toward downstream utility. As shown in Fig.~\ref{fig1}(c), the gap between fidelity-oriented demosaicking and SfP prevents further improvement in task performance, underscoring the need for a joint optimization paradigm that makes polarization reconstruction both physically accurate and task-aware.

We argue that these performance gaps stem from two fundamental flaws: ($i$) performing polarimetric task without essential demosaicking, and ($ii$) treating demosaicking and downstream tasks as isolated processes. To this end, we propose {\bf PolarAPP}, {\em the first framework to jointly optimize demosaicking and its downstream tasks}. PolarAPP incorporates a feature alignment mechanism optimized via meta-learning, which aligns the feature spaces of the demosaicking and task networks to produce task-adaptive demosaicking reconstructions and accurate task results. Then, inspired by equivalent imaging theory~\cite{chen2021equivariant,chen2022robust,chen2023imaging}, PolarAPP adopts an equivalent imaging transformation (EIT) prior that explicitly enables the network to focus on learning the intrinsic reconstruction mapping, rather than overfitting to dataset-specific artifacts. After joint optimization, we further refine the downstream model with a dedicated fine-tuning stage, leveraging the stable demosaicking front-end to further improve downstream accuracy under high-quality inputs. PolarAPP enables end-to-end learning from mosaic input to task-specific outputs by propagating task-level supervision throughout the entire pipeline. As shown in Fig.~\ref{fig1}(d), we validate the feasibility of jointly optimizing SfP and demosaicking, and further extend this approach to other tasks like DfP. Our contributions are summarized as follows:

\begin{itemize}
    \item We introduce PolarAPP, the first framework that jointly optimizes reconstruction and polarimetric applications, bridging low-level recovery with downstream polarization-based analysis tasks.
    \item We design a feature alignment mechanism that semantically links demosaicking and downstream task features through a customized meta-learning-based pipeline, guiding the reconstruction with task-aware feedback.
    \item We employ an EIT prior that decouples demosaicking from low-quality data artifacts, enabling physics-consistent and faithful polarization reconstruction under the constrained task-specific datasets.
    \item Extensive experiments demonstrate that PolarAPP outperforms conventional two-stage methods on demosaicking, SfP and DfP, validating the effectiveness of our joint, task-aligned design.
\end{itemize}

\section{Related Work}
\label{sec:rw}

\subsection{Polarization Image Demosaicking}

Early polarization demosaicking methods rely on interpolation~\cite{liu2020new, zhang2021polarizationT, morimatsu2021monochrome, xin2023demosaicking, lu2024a,Lu2024polarization} or optimization~\cite{wen2021sparse, qiu2021linear, dumoulin2022impact, luo2023sparse, luo2024learning} with hand-crafted priors (e.g., edge preservation, chrominance consistency), generally suffering from artifacts in complex scenes. Recent deep learning approaches leverage CNNs~\cite{sun2021color, li2022unsupervised, nguyen2022two}, GANs~\cite{guo2024attention}, or Transformers~\cite{li2025demosaicking, zhou2025pidsr} to learn end-to-end mappings, achieving superior reconstruction quality. However, these methods typically treat demosaicking as an isolated task, and thus ignore its intrinsic coupling with downstream tasks (e.g. SfP and DfP), limiting the overall performance.
%-------------------------------------------------------------------------
\subsection{Polarimetric Applications in SfP and DfP}

Polarimetric imaging supports a range of vision tasks such as SfP and DfP. SfP exploits the Fresnel reflection effect, where the degree and angle of polarization correlate with surface normals, enabling geometry recovery from a single polarized view. DfP utilizes the difference in polarization states between reflected and transmitted components, particularly near the Brewster angle~\cite{nayar1997separation, farid1999separating}, to separate specular reflections from diffuse layers. Both tasks have evolved from physics-based analytical formulations~\cite{nayar1997separation,farid1999separating,smith2016linear,smith2018height,logothetis2019differential,ngo2021surface,schechner2000polarization,deschaintre2021deep,ichikawa2021shape,chen2022perspective} to deep-learning approaches~\cite{ba2020deep,fukao2021polarimetric,dave2022pandora,lei2022shape,hwang2022sparse,shao2023transparent,li2024neisf,han2024nersp,lyu2024sfpuel,lyu2019reflection,lei2020polarized,yao2025polarfree}, which achieve improved robustness and reconstruction quality. Despite this progress, most learning-based approaches rely on polarization data constructed by rearranging DoFP sensor pixels into separate channels, discarding high-frequency details and bypassing the necessary demosaicking stage. This unrealistic preprocessing causes models to overlook mosaic-induced artifacts, leading to degraded accuracy in normal estimation and reflection removal. These limitations motivate a unified framework that couples polarization demosaicking with downstream polarimetric tasks, forming the basis of our proposed PolarAPP.

%-------------------------------------------------------------------------
\subsection{Meta-Learning in Vision}

Meta-learning develops algorithms to automatically fine-tune model parameters for specific tasks, demonstrating strong adaptability across domains. MAML~\cite{finn2017model, finn2019online, qin2023ground} learns task-agnostic initialization parameters that enable fast adaptation with limited data. Other approaches~\cite{ren2018learning, shu2019meta} focus on sample weighting, and they use a small validation set to identify informative samples under noise. Concurrently, some methods~\cite{zhao2023metafusion, bai2025task} explore learning parametric loss functions to better align training objectives with downstream tasks. However, no existing work applies meta-learning to polarimetric imaging, where sensor-level degradation critically impacts performance of downstream tasks. This gap motivates our joint PolarAPP, which leverages meta-learning to bridge the gap between low-level reconstruction and downstream shape and reflection estimation.

\section{Proposed Method}

\subsection{Overview}

The overall framework of PolarAPP is illustrated in Fig.~\ref{fig2}. During inference (Fig.~\ref{fig2}(a)), the pipeline takes a single CPFA raw image $\Rmat$ as input and outputs a task prediction $\Ymat$. We first apply Bayer demosaicking to obtain half-resolution color polarization images $\Imat^{\downarrow 2}$, which are fed into our color polarization demosaicker $\mathcal{D}$ to reconstruct full-resolution polarization images $\Imat = $ \{$\Imat_{0^\circ}$, $\Imat_{45^\circ}$, $\Imat_{90^\circ}$, $\Imat_{135^\circ}$\}. From $\Imat$, we compute physically interpretable polarization features~\cite{zhao2024polarization}, including Stokes components $\Smat$ ($\Smat_0$=$\frac{\Imat_{0^\circ} + \Imat_{45^\circ}+ \Imat_{90^\circ}+ \Imat_{135^\circ}}{2}$, $\Smat_1 = \Imat_{0^\circ} - \Imat_{90^\circ}$, $\Smat_2 = \Imat_{45^\circ} - \Imat_{135^\circ}$), $\text{DoLP} = \frac{\sqrt{\Smat_1^2 + \Smat_2^2}}{\Smat_0}$, $\text{AoP} = \frac{1}{2}\text{atan2}\left( {\Smat_2},{\Smat_1} \right)$, along with normalized spatial coordinates. These cues are concatenated channel-wise and fed into a task-specific network $\mathcal{T}$ to produce $\Ymat$.

{
\begin{figure}[t]
\centering
\includegraphics[width=0.99\textwidth]{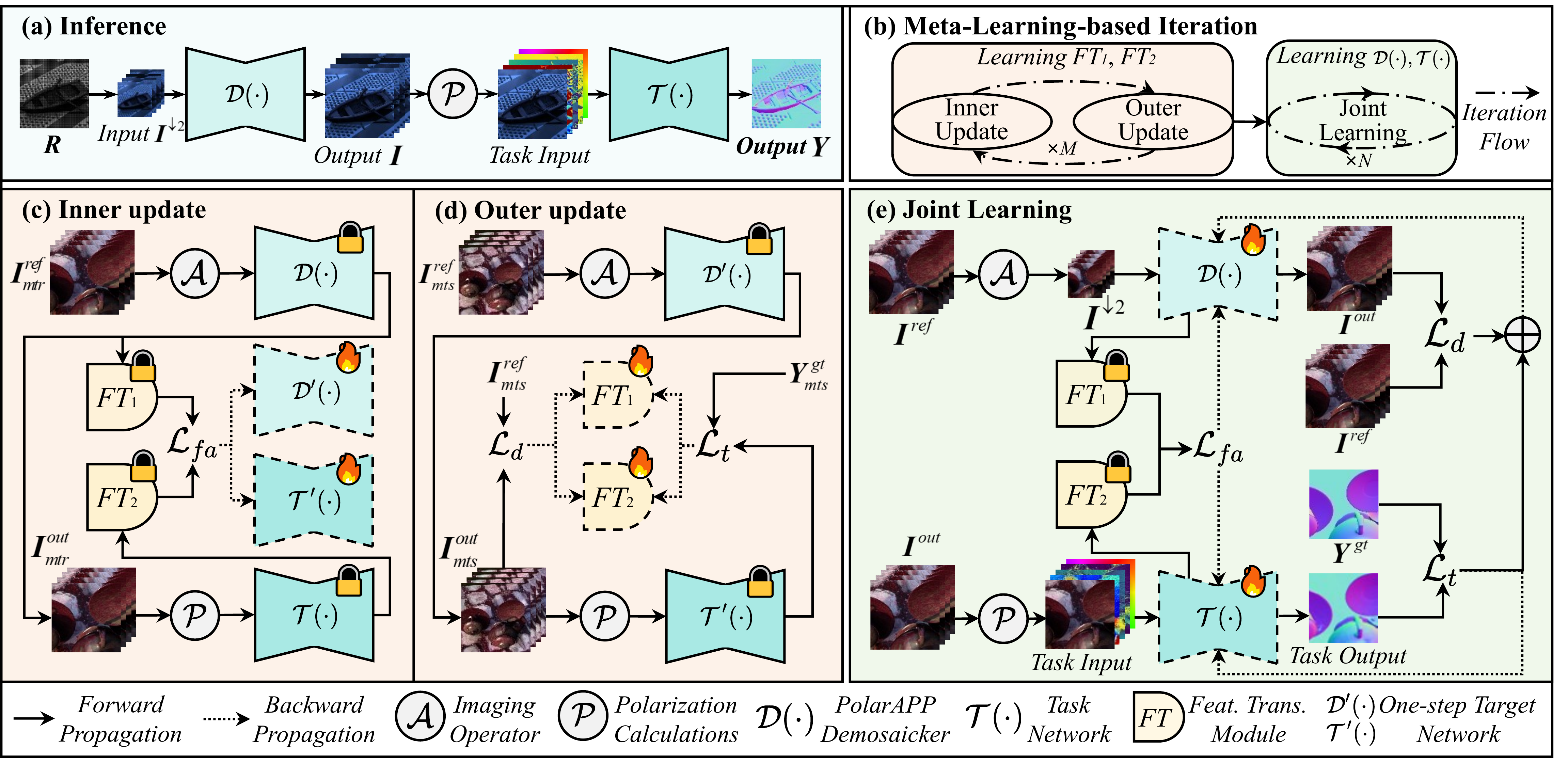}
\captionsetup{type=figure,skip=4pt}
\caption{The pipeline of PolarAPP. (a) Inference of PolarAPP. (b) Overview of the meta-learning-based iteration for joint training. (c) Inner update for updating demosaicking and task network under one gradient descent. (d) Outer update for updating the feature transform modules. (e) Joint demosaicking and task learning.}
\label{fig2}
\end{figure}}

Although inference resembles a standard two-stage pipeline, the key observation behind PolarAPP is that demosaicking and downstream performance are coupled. The task network
$\mathcal{T}$ is driven by physically derived polarization cues (DoLP/AoP), whose fidelity is directly determined by the demosaicked polarization images produced by $\mathcal{D}$. Consequently, a task-agnostic demosaicker optimized solely for reconstruction fidelity may preserve appearance yet distort subtle polarization structures that are critical to SfP/DfP, motivating joint learning to make $\mathcal{D}$ explicitly task-aware. However, na\"ive end-to-end joint training with demosaicking loss $\mathcal{L}_d$ and task loss $\mathcal{L}_t$ is prone to objective mismatch, where reconstruction fidelity and task objectives may induce conflicting gradients and the interaction between the two networks is mediated only through intermediate images. PolarAPP instead introduces feature alignment and meta-learns an alignment metric via a bi-level optimization, so that the alignment objective is updated only when it improves the end objectives measured by $\mathcal{L}_d$ and $\mathcal{L}_t$. Importantly, the alignment modules are used only during training and are discarded at inference, introducing zero additional inference overhead.

%-------------------------------------------------------------------------
\subsection{Loss Function}

Before detailing the meta-learning training strategy, we define the loss functions used in the PolarAPP. These include standard reconstruction and task-specific losses, as well as a feature alignment loss that enables task-aware optimization.

\noindent \textbf{Demosaicking loss}. $\mathcal{D}$ is trained to reconstruct full-resolution polarization images, from which the $\Smat$, $\text{DoLP}$, and $\text{AoP}$ are derived. To ensure high-fidelity reconstruction, we adopt a composite loss function $\mathcal{L}_d$, which jointly constrains $\Imat$, $\Smat$ and $\text{DoLP}$ through $\ell_1$ and gradient losses inspired by PIDSR~\cite{zhou2025pidsr}. AoP is supervised with $\ell_1$ loss on wrapped angular differences to respect periodicity.

\noindent \textbf{Task-Specific loss}. $\mathcal{T}$ is trained with a task-specific loss $\mathcal{L}_t$, designed according to the downstream application. For SfP, we use cosine similarity and mean angular error losses to supervise surface normal estimation. For DfP, we follow the training protocol of PolarFree~\cite{yao2025polarfree}, incorporating pixel, VGG, total variation, phase, and diffusion prior losses.

\noindent \textbf{Feature Alignment loss}. To connect demosaicking and downstream task for mutual promotion, we introduce a feature alignment loss $\mathcal{L}_{fa}$, which operates through a meta-learned feature metric. Unlike $\mathcal{L}_{d}$ and $\mathcal{L}_{t}$, which provide direct supervision, $\mathcal{L}_{fa}$ encourages the demosaicking features to be task-ready and the task features to be reconstruction-aware. Specifically, in the meta-training loop, two feature transform modules ($FT$, Each FT consists of three CNN blocks) extract features $\Fmat_{dj}$ and $\Fmat_{tj}$ (where $j \in \{ 1,2,3 \}$ indexes the feature hierarchy) from the output of $\mathcal{D}$ and $\mathcal{T}$ for calculating $\mathcal{L}_{fa}$. $\mathcal{L}_{fa}$ is defined as $\mathcal{L}_{fa} = 3-\sum_{j = 1}^3\frac{\langle FT_1(\Fmat_{dj}), FT_2(\Fmat_{tj}) \rangle}{\left\| FT_1(\Fmat_{dj}) \right \|_2 \cdot \left\| FT_2(\Fmat_{tj}) \right \|_2}$, where the FT modules are meta-trained in the outer loop to make $\mathcal{L}_{fa}$ a meaningful signal for joint optimization. The total loss $\mathcal{L}$ is denoted as $\mathcal{L}_{d}+\lambda_{t}\mathcal{L}_{t}+\lambda_{fa}\mathcal{L}_{fa}$, where $\lambda_{t}$ and $\lambda_{fa}$ balance the trade-off.

%-------------------------------------------------------------------------
\subsection{Meta-Learning-based Feature Alignment}

As discussed in Sec.~3.1, na\"ive joint training of $\mathcal{D}$ and $\mathcal{T}$ with $\mathcal{L}_{d}$+$\mathcal{L}_{t}$ can suffer from objective mismatch and limited cross-network communication. We therefore introduce two FT modules \{$FT_1$, $FT_2$\} to align intermediate representations and enable feature-level information exchange in a latent metric space. However, learning a useful shared metric space is non-trivial: directly constraining the original feature spaces provides weak or misleading gradients due to representation discrepancy, and a fixed alignment loss can admit degenerate solutions (e.g., collapsed projections that minimize $\mathcal{L}_{fa}$ without improving $\mathcal{L}_d$ and $\mathcal{L}_t$). To address this, we meta-learn the alignment metric via a bi-level scheme (Fig.~\ref{fig2}(c-d)): the inner update uses $\mathcal{L}_{fa}$ to obtain tentative parameters ($\mathcal{D}'$, $\mathcal{T}'$), and the outer update optimizes \{$FT_1$, $FT_2$\} only through the resulting decreases in $\mathcal{L}_{d}$ and $\mathcal{L}_{t}$ on a meta-test split. This ties the alignment space to task and reconstruction objectives, yielding a goal-directed $\mathcal{L}_{fa}$ that stabilizes joint learning of $\mathcal{D}$ and $\mathcal{T}$.

\noindent \textbf{Inner Update}. Given the meta-train set $\{\Imat_{mtr}^{ref},\Ymat_{mtr}^{gt}\}$, we form the input to $\mathcal{D}$ by applying the imaging operator $\mathcal{A}$ to the reference color polarization image $\Imat_{mtr}^{ref}$, where $\mathcal{A}$ denotes CPFA down-sampling and Bayer interpolation. Here, $\Imat^{ref}$ denotes the color polarization image obtained by regrouping and Bayer demosaicking into half resolution (relative to real-captured sensor resolution), which serves as the supervision target of $\mathcal{L}_d$ in the meta-learning loop. In the inner step, we use $\mathcal{L}_{fa}$ to perform a probe update that assesses the quality of the current alignment metric. Specifically, both $\mathcal{D}$ and $\mathcal{T}$ take one gradient descent step guided by $\mathcal{L}_{fa}$ computed from the meta-features $\Fmat_{dj}^m$ and $\Fmat_{tj}^m$:
\begin{align}
    \theta_{\mathcal{D}'} 
    &= \textstyle \theta_{\mathcal{D}} - \beta_{\mathcal{D}'} \sum_{j=1}^{3} 
    \frac{\partial \mathcal{L}_{fa}(FT_1(\Fmat_{dj}^m), FT_2(\Fmat_{tj}^m))}{\partial \theta_{\mathcal{D}}},
    \label{eq:1} \\
    \theta_{\mathcal{T}'} 
    &= \textstyle \theta_{\mathcal{T}} - \beta_{\mathcal{T}'} \sum_{j=1}^{3} 
    \frac{\partial \mathcal{L}_{fa}(FT_1(\Fmat_{dj}^m), FT_2(\Fmat_{tj}^m))}{\partial \theta_{\mathcal{T}}},
    \label{eq:2}
\end{align}
where $\beta_{\mathcal{D}'}$\&$\beta_{\mathcal{T}'}$ are the inner-step learning rates and $\Imat_{mtr}^{out}$ the inner-step demosaicked output. The original parameters ($\theta_{\mathcal{D}}$, $\theta_{\mathcal{T}}$) of $(\mathcal{D},\mathcal{T})$ remain unchanged, and the temporary $(\mathcal{D}',\mathcal{T}')$ are created only to preserve the computational graphs required for optimizing the \{$FT_1$, $FT_2$\} in the outer update.

\begin{figure}[t]
    \centering
    \includegraphics[width=0.99\textwidth]{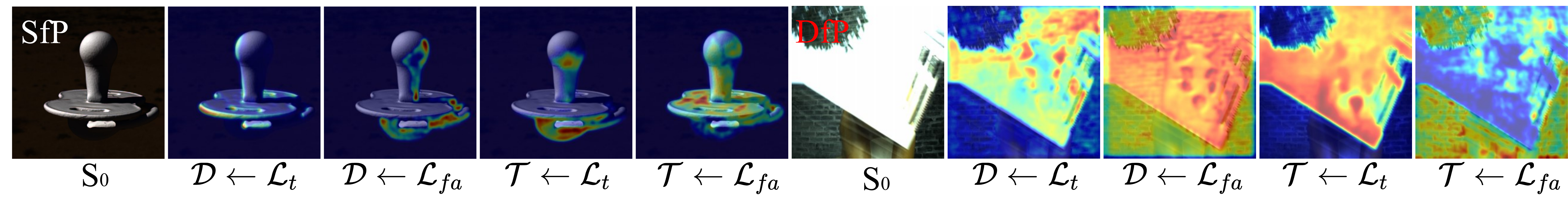}
    \captionsetup{type=figure,skip=0pt}
    \caption{Grad-CAM visual comparison of $\mathcal{L}_t$ and $\mathcal{L}_{fa}$ on SfP and DfP. It can be seen that $\mathcal{L}_{fa}$ yields more distributed, structure-aware responses for both $\mathcal{D}$ and $\mathcal{T}$.}
    \label{fig3}
\end{figure}

\noindent \textbf{Outer Update}. Using the meta-test set \{$\Imat_{mts}^{ref},\Ymat_{mts}^{gt}$\}, we update $FT$ parameters $\theta_{n}$ ($n$=$1,2$) under supervision from the ($\mathcal{L}_{d}$, $\mathcal{L}_{t}$) computed with (${\mathcal{D}'}$, ${\mathcal{T}'}$):
\begin{equation}
    \begin{split}
    \theta_{n} 
    % &= \theta_{n}-\beta_{n} (\nabla_{FT_{n}} \mathcal{L}_{d}(\Imat_{mts},\Imat_{mts}^{ref})+ \nabla_{FT_{n}} \mathcal{L}_{t}(\Ymat_{mts},\Ymat_{mts}^{gt})) \\
    & \textstyle \leftarrow \theta_{n} - \beta_{n} \frac{\partial \mathcal{L}_{d}(\mathcal{D}’(\mathcal{A}(\Imat_{mts}^{ref});\theta_{\mathcal{D}’}), \Imat_{mts}^{ref})}{\partial \theta_{n}} - \beta_{n} \frac{\partial \mathcal{L}_{t}(\mathcal{T}’(\Imat_{mts};\theta_{\mathcal{T}’}), \Ymat_{mts}^{gt})}{\partial \theta_{n}} \\
    &= \textstyle  \theta_{n} + \beta_{n} \cdot \beta_{\mathcal{D}’} \frac{\partial \mathcal{L}_{d}(\mathcal{D}'(\mathcal{A}(\Imat_{mts}^{ref});\theta_{\mathcal{D}'}),\Imat_{mts}^{ref})}{\partial \theta_{\mathcal{D}'}} \cdot  \frac{\partial^2 \mathcal{L}_{fa}}{\partial \theta_{n}\partial \theta_{\mathcal{D}}} \\
    & \qquad \textstyle  + \beta_{n} \cdot \beta_{\mathcal{T}’} \frac{\partial \mathcal{L}_{t}(\mathcal{T}'(\Imat_{mts};\theta_{\mathcal{T}'}),\Ymat_{mts}^{gt})}{\partial \theta_{\mathcal{T}'}} \cdot  \frac{\partial^2 \mathcal{L}_{fa}}{\partial \theta_{n}\partial \theta_{\mathcal{T}}},
    \end{split}
    \label{eq:3}
\end{equation}
where $\beta_n$ is the outer-step learning rate and $\Imat_{mts}^{out}$ is the outer-step demosaicked output. Backpropagating through Eq.~\eqref{eq:1}--\eqref{eq:2} yields second-order gradients that adjust \{$FT_1$, $FT_2$\} so that an $\mathcal{L}_{fa}$-induced inner update leads to lower $\mathcal{L}_d$ and $\mathcal{L}_t$ on the meta-test split.

The bi-level optimization makes $\mathcal{L}_{fa}$ goal-directed: the alignment metric is updated only when it improves the end objectives measured by $\mathcal{L}_d$ and $\mathcal{L}_t$. This prevents degenerate alignment and yields an alignment signal that can be reused in the subsequent joint learning stage. In addition, we visualize gradient-based attribution for $\mathcal{D}$ and $\mathcal{T}$ under $\mathcal{L}_t$ and $\mathcal{L}_{fa}$ in Fig.~\ref{fig3}. For SfP, $\mathcal{L}_{fa}$ emphasizes low-intensity regions in $\Smat_0$, complementing task supervision. For DfP, $\mathcal{L}_{fa}$ highlights object in strong reflection regions on $\mathcal{D}$ compared to $\mathcal{L}_{t}$; correspondingly, $\mathcal{L}_{t}$ drives $\mathcal{T}$'s gradients on dominant reflection regions, whereas $\mathcal{L}_{fa}$ distributes them to broader structural areas, encouraging structure-preserving predictions. Overall, the visualization indicates that the meta-learned alignment provides complementary training-time regularization with zero inference overhead.

%-------------------------------------------------------------------------
\subsection{Joint Learning}

Using the meta-learned alignment metric, we perform joint learning of the $\mathcal{D}$ and $\mathcal{T}$, as illustrated in Fig.~\ref{fig2}(e). Concretely, we adopt an epoch-wise alternating schedule: within each epoch, we first update $\{FT_1,FT_2\}$ via the bi-level inner/outer steps, and then freeze $\{FT_1,FT_2\}$ and update $\mathcal{D}$ and $\mathcal{T}$ using standard end-to-end backpropagation. 
This design ensures that meta-learning is responsible only for learning the alignment metric, while joint learning performs the actual task-aware optimization of $\mathcal{D}$ and $\mathcal{T}$.

During joint learning, we compute $\mathcal{L}_{fa}$ on the current feature hierarchies $\{\Fmat_{dj}\}$ and $\{\Fmat_{tj}\}$ extracted from $\mathcal{D}$ and $\mathcal{T}$ (with $j\in\{1,2,3\}$), and backpropagate it jointly with $\mathcal{L}_d$ and $\mathcal{L}_t$. The update of $\mathcal{D}$ is:
\begin{equation}
    \begin{split}
    \theta_{\mathcal{D}}
    & \textstyle \leftarrow \theta_{\mathcal{D}}-\beta_{{\mathcal{D}}} ( \frac{\partial \mathcal{L}_{d}(\Imat, \Imat^{ref}) + \lambda_{t}\partial \mathcal{L}_{t}(\Ymat, \Ymat^{gt})+\lambda_{fa}\sum_{j=1}^{3}\partial \mathcal{L}_{fa}(FT_1(\Fmat_{dj}), FT_2(\Fmat_{tj}))} {\partial \theta_{\mathcal{D}}}),
    \end{split}
    \label{eq:4}
\end{equation}
and the ${\mathcal{T}}$ is updated as:
\begin{equation}
    \theta_{\mathcal{T}}
    \textstyle \leftarrow \theta_{\mathcal{T}}-\beta_{{\mathcal{T}}} ( \frac{\lambda_{t}\partial \mathcal{L}_{t}(\Ymat, \Ymat^{gt})+\lambda_{fa}\sum_{j=1}^{3}\partial \mathcal{L}_{fa} (FT_1(\Fmat_{dj}), FT_2(\Fmat_{tj}))} {\partial \theta_{\mathcal{T}}}).
    \label{eq:5}
\end{equation}
Here, gradients of $\mathcal{L}_{fa}$ flow through the fixed transforms $\{FT_1,FT_2\}$ into both networks, enabling feature-level communication while keeping inference unchanged (the transforms are used only for training). Overall, this stage uses the meta-learned alignment objective to bias demosaicking toward task-relevant structures and to provide the task network with reconstruction-aware context, so that improvements in $\mathcal{D}$ translate more directly into downstream performance.

%-------------------------------------------------------------------------
{
\begin{figure}[t]
    \centering
    \includegraphics[width=0.99\textwidth]{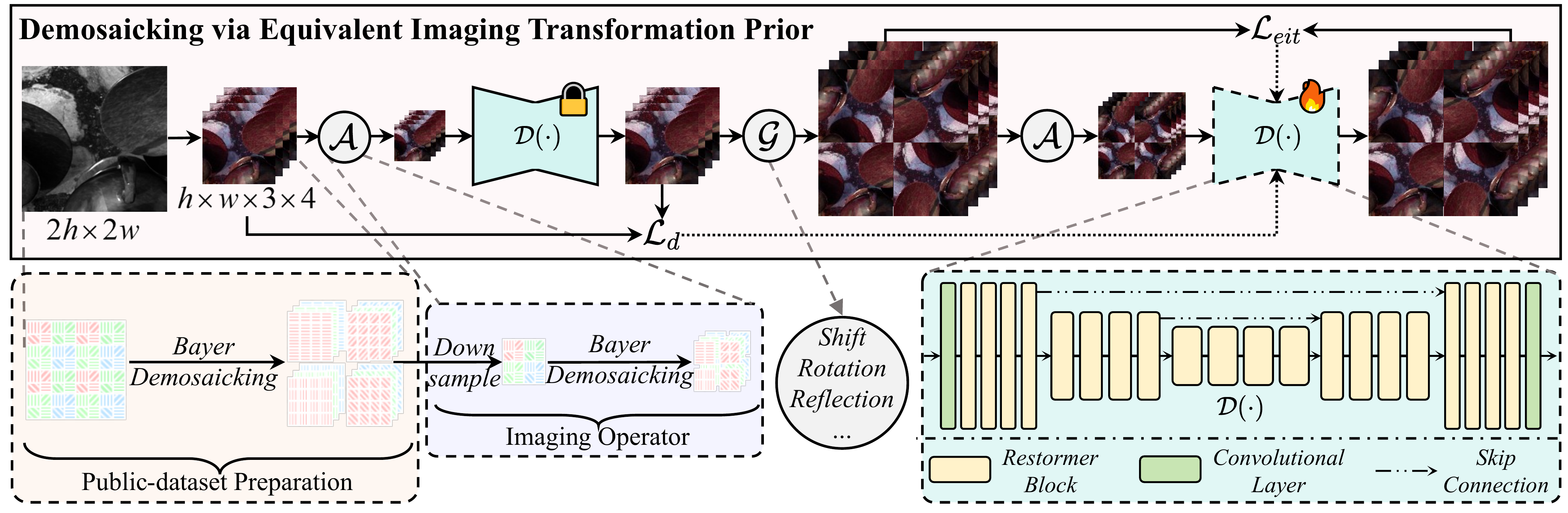}
    \captionsetup{type=figure,skip=4pt}
    \caption{Learning demosaicking based on equivalent imaging transformation prior.}
    \label{fig4}
\end{figure}}

\subsection{Learning Demosaicking via EIT prior}
Standard demosaicking datasets lack corresponding task GT, while polarization-task datasets (e.g., SfP or DfP) contain four polarization angles but often suffer from low image quality due to extraction from single-frame CPFA captures. Although we introduce $\mathcal{A}$ to construct paired data on task-specific datasets for training, its inherent low quality prevent reconstructions from high performance when $\mathcal{D}$ is only supervised by $\mathcal{L}_d$. To overcome this limitation, we first introduce the EIT prior for demosaicking, a self-supervised constraint that incorporates the imaging process. Since imaging operator is fully known and deterministic, EIT enables learning imaging-aware representation from limited observation by leveraging transformation consistency, thereby overcoming the performance bottleneck of supervised learning on constrained task-specific datasets.

The training of demosaicking is shown in Fig.~\ref{fig4}, where $\mathcal{D}$ is a Restormer-inspired UNet~\cite{ronneberger2015u,zamir2022restormer}, and $\mathcal{G}$ denotes a set of geometric transformations $T_g$. In practice, $T_g$ includes one random non-zero integer translation, one $10^{\circ}$ rotation, and one horizontal flip, all applied to the reconstructed full-resolution polarization image before re-sampling by $\mathcal{A}$. The equivariant demosaicking hypothesis asserts that, for any $T_g \in \mathcal{G}$:
\begin{equation}
    \mathcal{D}(\mathcal{A}(T_g \Imat^{ref})) = T_g \mathcal{D} (\mathcal{A}(\Imat^{ref})).
    \label{eq:6}
\end{equation}
This means that applying a geometric transformation in the reference polarization-image domain and then demosaicking should be consistent with demosaicking first and then transforming the reconstruction. This equivariance constraint provides a self-consistency condition that can be enforced without additional labeled data. Specifically, given a reference polarization image $\Imat^{ref}$, we first obtain its reconstruction:
\begin{equation}
    \Imat^{rec}=\mathcal{D}(\mathcal{A}(\Imat^{ref})).
    \label{eq:7}
\end{equation}
We then apply $T_g$ in the reconstructed polarization-image domain and re-sample it using the fixed imaging operator $\mathcal{A}$ to synthesize a valid CPFA-consistent observation. The EIT loss is defined as:
\begin{equation}
    \mathcal{L}_{eit}
    =
    \mathbb{E}_{\Imat^{ref},g}
    \left[
    \mathcal{L}_{d}
    \left(
    \mathcal{D}(\mathcal{A}(T_g \Imat^{rec})),
    T_g\mathcal{D}(\mathcal{A}(\Imat^{ref}))
    \right)
    \right].
    \label{eq:8}
\end{equation}

\begin{figure}[t]
    \centering
    \includegraphics[width=0.95\columnwidth]{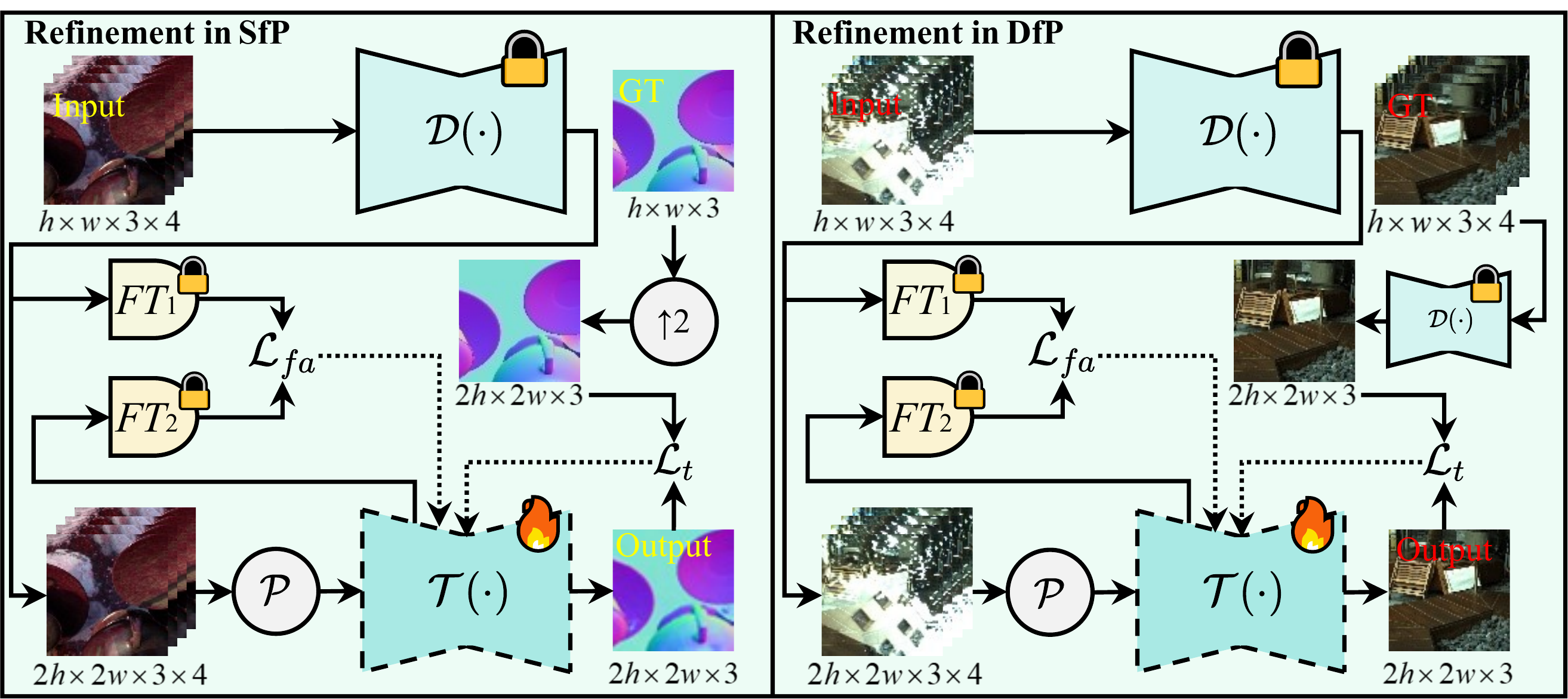}
    \captionsetup{type=figure,skip=4pt}
    \caption{Overview of the task-refinement stage for SfP and DfP. $\mathcal{D}$ remains fixed while $\mathcal{T}$ is optimized with clean polarization inputs to enhance task accuracy.}
    \label{fig5}
\end{figure}

\begin{algorithm}[t]
\caption{PolarAPP Training Procedure}
\label{alg:1}
\KwIn{Training set $\{(\Imat^{ref}, \Ymat^{gt})\}$}
\KwOut{Learned parameters $\theta_{\mathcal{D}}, \theta_{\mathcal{T}}, \theta_{1}, \theta_{2}$}

Initialize $\theta_{\mathcal{D}}, \theta_{\mathcal{T}}, \theta_{1}, \theta_{2}$\;

\While{PolarAPP training}{
    \textit{// Meta-learning of the feature alignment metric}\;
    \While{feature-alignment meta-learning}{
        Sample $(\Imat_{mtr}^{ref}, \Ymat_{mtr}^{gt})$\;
        Obtain $(\Fmat_d^m, \Fmat_t^m)$ and update $\theta_{\mathcal{D}'}, \theta_{\mathcal{T}'}$ by Eq.~\eqref{eq:1}--Eq.~\eqref{eq:2}\;
        Sample $(\Imat_{mts}^{ref}, \Ymat_{mts}^{gt})$\;
        Obtain $(\Imat_{mts}, \Ymat_{mts})$ and update $\theta_1, \theta_2$ by Eq.~\eqref{eq:3}\;
    }

    \textit{// Joint learning of demosaicking and downstream task}\;
    \While{joint demosaicking-task learning}{
        Sample $(\Imat^{ref}, \Ymat^{gt})$\;
        Obtain $(\Imat, \Ymat, \Fmat_d, \Fmat_t)$ and update $\theta_{\mathcal{D}}, \theta_{\mathcal{T}}$ by Eq.~\eqref{eq:4}--Eq.~\eqref{eq:5}\;
    }

    \textit{// Task-network refinement with frozen demosaicker}\;
    \While{task-network refinement}{
        Sample $(\Imat^{ref}, \Ymat^{gt})$ without applying $\mathcal{A}$\;
        Obtain $(\Imat^{\uparrow 2}, \Ymat^{\uparrow 2}, \Fmat_d^{\uparrow 2}, \Fmat_t^{\uparrow 2})$\;
        Obtain $(\Imat_{ref}^{\uparrow 2}, \Ymat_{gt}^{\uparrow 2})$ and update $\theta_{\mathcal{T}}$ by Eq.~\eqref{eq:5}\;
    }
}
\end{algorithm}

\noindent In this way, transformations are applied before re-sampling by $\mathcal{A}$, so the CPFA or Bayer sampling layout remains valid and the generated inputs are still consistent with the known imaging process. $\mathcal{L}_{eit}$ regularizes $\mathcal{D}$ to produce consistent polarization reconstructions under different transformed observations. The demosaicking objective becomes $\mathcal{L}_d \leftarrow \mathcal{L}_d+\mathcal{L}_{eit}$.

%-------------------------------------------------------------------------
\subsection{Task Refinement and Training}
\noindent{\bf Task Refinement:}
As mentioned earlier, joint training begins by generating low-quality inputs paired with reference images through $\mathcal{A}$. Although $\mathcal{D}$ performs well with the aid of the EIT prior, the compression introduced by $\mathcal{A}$ limits $\mathcal{D}$’s output quality during training, thereby hindering the improvement of $\mathcal{T}$. Therefore, we perform a refinement stage after the joint learning, where $\mathcal{A}$ is removed to fully unleash $\mathcal{D}$'s restoration capability. Notably, $\mathcal{D}$ is frozen in this stage, and $\mathcal{T}$ is optimized using $\mathcal{L}_t$ and $\mathcal{L}_{fa}$, as shown in Fig.~\ref{fig5}. Removing $\mathcal{A}$ increases the task output resolution, thus requiring upsampling of the corresponding task GT for supervision. Specially, we use bilinear interpolation to upsample the SfP normals GT purely for scale alignment. The $2\times$ reflection-free supervision for DfP is obtained by applying the \emph{frozen} $\mathcal{D}$ to generate full-resolution pseudo references (used only for supervision, without updating $\mathcal{D}$). This refinement brings two benefits: ($i$) $\mathcal{D}$ processes full-resolution inputs, yielding higher-quality demosaicking images for downstream training; and ($ii$) aligns training with inference conditions, reducing the resolution gap. Overall, it consolidates joint optimization and improves task accuracy under realistic inputs.

%-------------------------------------------------------------------------

\noindent{\bf Training:}
The complete training procedure is summarized in Alg.~\ref{alg:1}. Each epoch alternates among ($i$) meta-learning (inner/outer updates to refine $FT_1$, $FT_2$), ($ii$) joint learning (updating $\mathcal{D}$, $\mathcal{T}$ using $\mathcal{L}_d$, $\mathcal{L}_t$, $\mathcal{L}_{fa}$), and ($iii$) task refinement. For the SfP task, $\mathcal{T}$ adopts an architecture similar to $\mathcal{D}$; for DfP, we integrate the PolarFree framework to provide a diffusion prior.

\section{Experiment}

\subsection{Experimental Setting}
In training, the number of epochs $L$ and the meta-learning iterations $M$ are set to 100 and 200, respectively. The joint learning iterations $N$ depends on the training set size. The batch size is fixed to 2, and all models in different training stages are optimized using Adam with a learning rate of $5 \times 10^{-5}$. The hyperparameters $\lambda_t$ are set 20 for balanced scaling, while $\lambda_{fa}$ is chosen based on ablation study. All experiments are conducted on servers equipped with NVIDIA RTX 4090 GPUs.

\noindent \textbf{Dataset Preparation}. PolarAPP is evaluated on polarization tasks including SfP and DfP. For SfP, we use the SfPUEL dataset~\cite{lyu2024sfpuel}, where 19,800 of 20,000 image groups are used for training and the remainder, together with several real-captured scenes, are used for testing. For DfP, 6,312 of 6,500 image groups from the PolarRGB dataset~\cite{yao2025polarfree} are used for training and the rest for testing. In addition, the Qiu dataset~\cite{qiu2021linear} and several DoFP raw images are included as supplementary benchmarks for evaluating polarization demosaicking quality.

\noindent \textbf{Comparison methods}. For demosaicking evaluation, we compare PolarAPP with SOTA methods DCPM~\cite{li2025demosaicking} and PIDSR~\cite{zhou2025pidsr}, and they are re-trained on the same dataset as PolarAPP's. The $\Smat_0$ and $\text{DoLP}$ are evaluated using PSNR and SSIM, while the $\text{AoP}$ is measured by Mean Angular Error (MAE). For SfP comparison, SfPW~\cite{lei2022shape} and SfPUEL~\cite{lyu2024sfpuel} are applied either to demosaicked images or to direct inputs, forming $4 \times 3$ result combinations. Since SfP primarily concerns normal-direction accuracy rather than image-level appearance, the predicted normal maps are evaluated by angular-accuracy thresholds ($11.25^\circ$, $22.5^\circ$, $30.0^\circ$) and angle error metrics (mean, median, RMSE). For DfP comparison, PolarFree~\cite{yao2025polarfree} is paired with different upstream modules, producing another $4 \times 2$ set of results. Because DfP jointly requires effective reflection removal and detail preservation, the de-reflected outputs are assessed using PSNR, SSIM, LPIPS~\cite{zhang2018unreasonable} and MUSIQ~\cite{ke2021musiq} metrics.

\begin{figure}[t]
\centering
\begin{minipage}[t]{0.40\textwidth}
    \null
    \setlength{\belowcaptionskip}{3pt}
    \captionof{table}{Metrics comparisons on demosaicking.}
    \label{tab:1}
    \small
    \centering
    \setlength{\tabcolsep}{4pt}
    \renewcommand{\arraystretch}{1.2}
    \resizebox{\linewidth}{!}{
    \begin{tabular}{l|ccc}
    \hline
    Methods                   & DCPM  & PIDSR & Ours           \\ \hline
                              & \multicolumn{3}{c}{$S_0$}         \\ \cline{2-4}
    PSNR(dB)                  & 42.98 & 43.07 & \textbf{43.12} \\
    SSIM                      & 0.976 & 0.973 & \textbf{0.982} \\ \cline{2-4}
                              & \multicolumn{3}{c}{DoLP}          \\ \cline{2-4}
    PSNR(dB)                  & 37.92 & 37.86 & \textbf{38.10} \\
    SSIM                      & 0.918 & 0.925 & \textbf{0.928} \\ \cline{2-4}
                              & \multicolumn{3}{c}{AoP}           \\ \cline{2-4}
    MAE($^\circ$)             & 6.456 & 5.587 & \textbf{5.444} \\ \hline
    \end{tabular}}
\end{minipage}\hfill
\begin{minipage}[t]{0.58\textwidth}
    \null
    \centering
    \includegraphics[width=\linewidth]{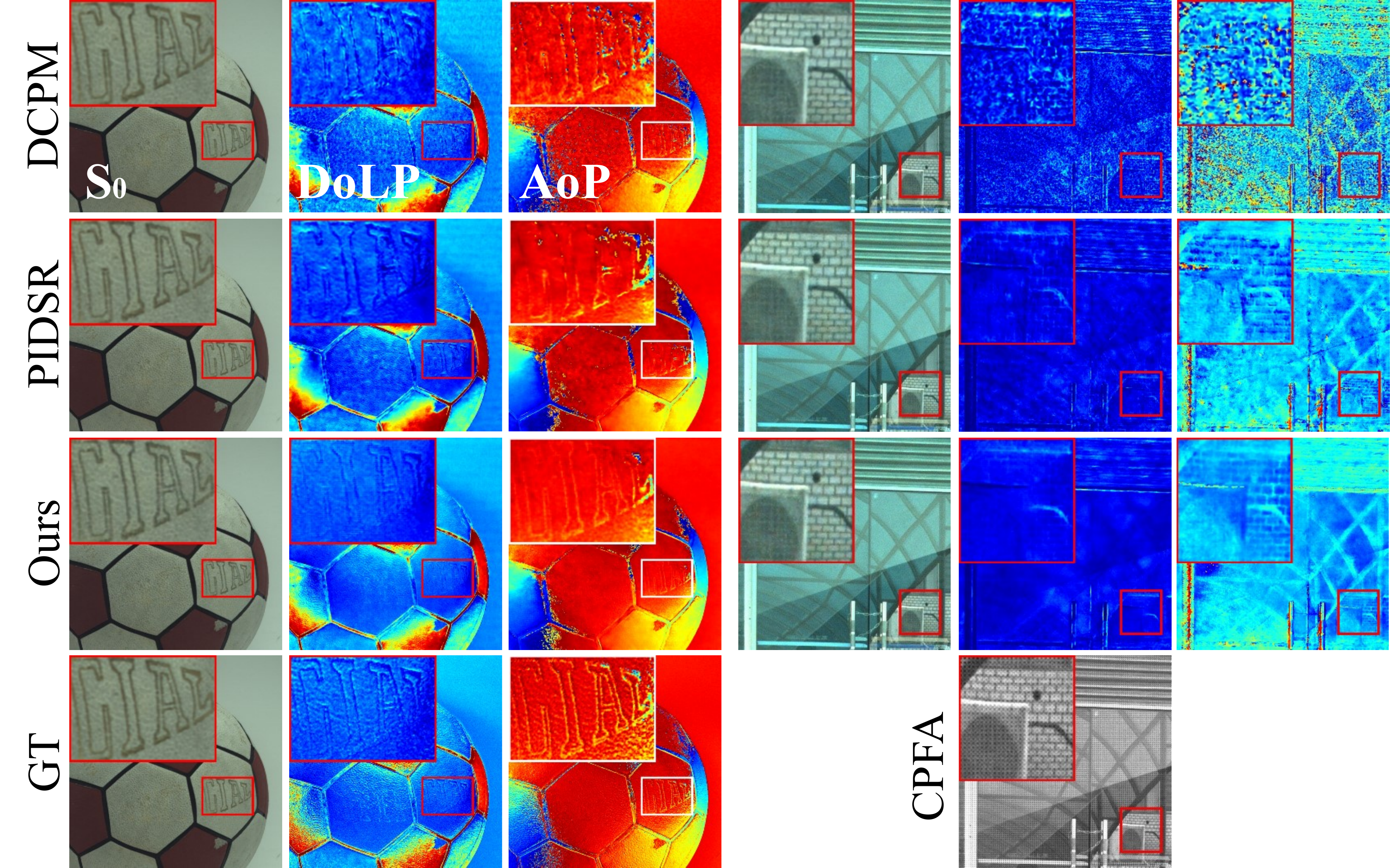}
    \captionsetup{type=figure,skip=-5pt}
    \captionof{figure}{Visual comparison in demosaicking.}
    \label{fig6}
\end{minipage}
\end{figure}

\begin{table}[t]
\setlength{\belowcaptionskip}{3pt}
\caption{Quantitative task comparisons. PIDSR+ denotes retraining the corresponding task networks, SfPUEL for SfP and PolarFree for DfP, on outputs from a fixed PIDSR demosaicker.}
\label{tab:2}
\centering
\setlength{\tabcolsep}{3pt}
\renewcommand{\arraystretch}{1.0}
\resizebox{\textwidth}{!}{
\begin{tabular}{c|cccccc|c|cccc}
\hline
\rule{0pt}{10pt}\multirow{2}{*}{SfP} 
& \multicolumn{3}{c}{Accuracy $\uparrow$ (\%)} 
& \multicolumn{3}{c|}{Error $\downarrow$ ($^\circ$)} 
& \multirow{2}{*}{DfP} 
& \multicolumn{3}{c}{$w$ Ref.} 
& \multicolumn{1}{c}{$w/o$ Ref.} \\[2pt]

\rule{0pt}{10pt} 
& 11.25$^\circ$ & 22.5$^\circ$ & 30$^\circ$
& Mean & Med. & RMSE
& 
& PSNR$\uparrow$ & SSIM$\uparrow$ & LPIPS$\downarrow$ & MUSIQ$\uparrow$ \\[1pt]
\hline
\addlinespace[1pt]

SfPUEL    
& 70.11 & 91.15 & 95.02 
& 10.58 & 7.59 & 15.30 
& PolarFree       
& 22.44 & 0.868 & 0.132 & 60.31 \\

PIDSR+  
& 78.12 & 90.73 & 95.55 
& 6.78 & 4.88 & 11.45 
& PIDSR+
& 22.41 & 0.859 & \textbf{0.124} & 60.79 \\

Ours          
& \textbf{92.37} & \textbf{97.45} & \textbf{98.48} 
& \textbf{3.57} & \textbf{2.71} & \textbf{8.64} 
& Ours
& \textbf{22.90} & \textbf{0.871} & 0.129 & \textbf{61.29} \\
\hline
\end{tabular}
}
\end{table}

\subsection{Demosaicking Comparison}
The quantitative results on the Qiu dataset are summarized in Tab.~\ref{tab:1}. PolarAPP delivers demosaicking performance that is competitive with SOTA methods while maintaining the stability required for downstream tasks--a balance that purely fidelity-driven demosaickers do not achieve. Visual comparisons on both Qiu and real-world raw data (Fig.~\ref{fig6}) show that PolarAPP preserves edge structure and polarization patterns more reliably than DCPM and PIDSR, which often produce blurred or unstable DoLP/AoP maps. In real scenes, PolarAPP reconstructs polarization cues that remain visually coherent and physically plausible, highlighting its stronger generalization under unconstrained conditions. Additional qualitative results are provided in the supplementary material (SM).

\begin{figure}[t]
\centering
    \includegraphics[width=.99\textwidth]{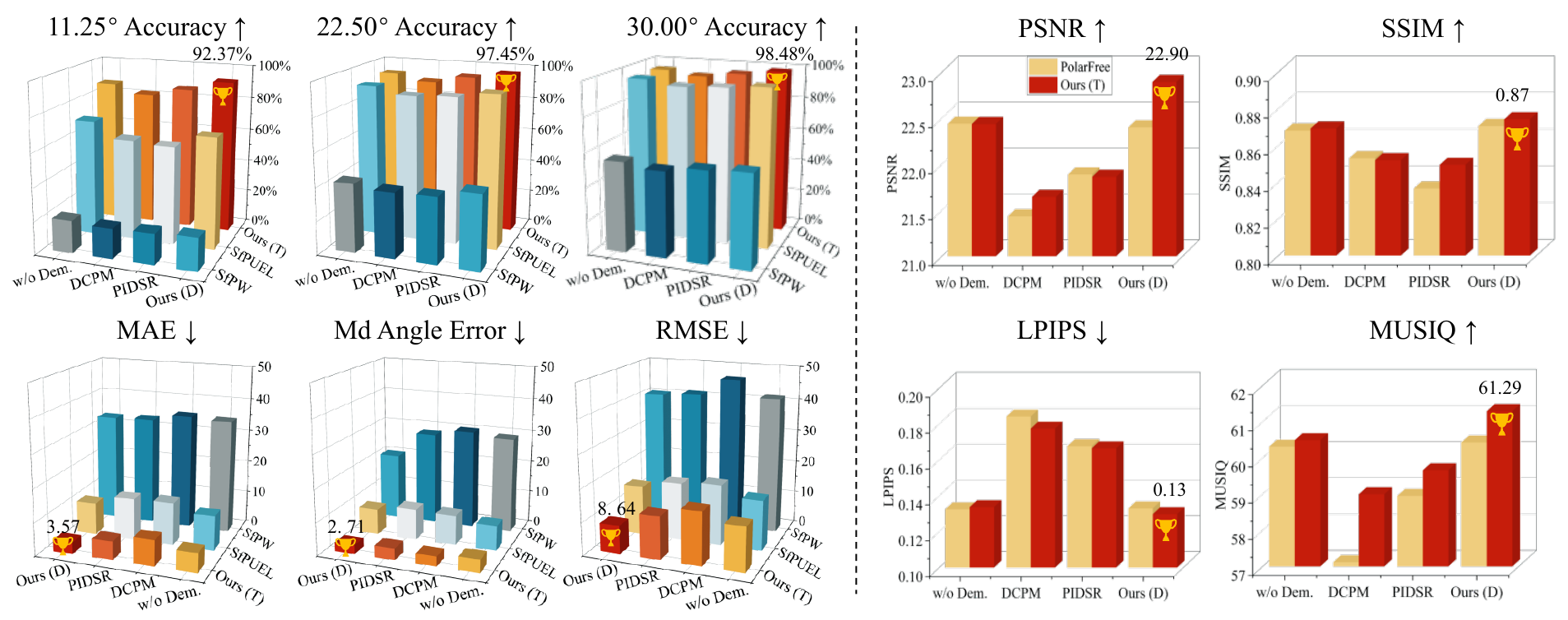}
    \captionsetup{type=figure,skip=4pt}
    \caption{Combination landscape in SfP (left) and DfP (right). Each column denotes combination of one demosaicker and one task method, serving as a practical reference for module compatibility. Complete PolarAPP shows the best metrics on both tasks.}
\label{fig7}
\end{figure}

{
\begin{figure}[t]
\centering
    \includegraphics[width=0.99\textwidth]{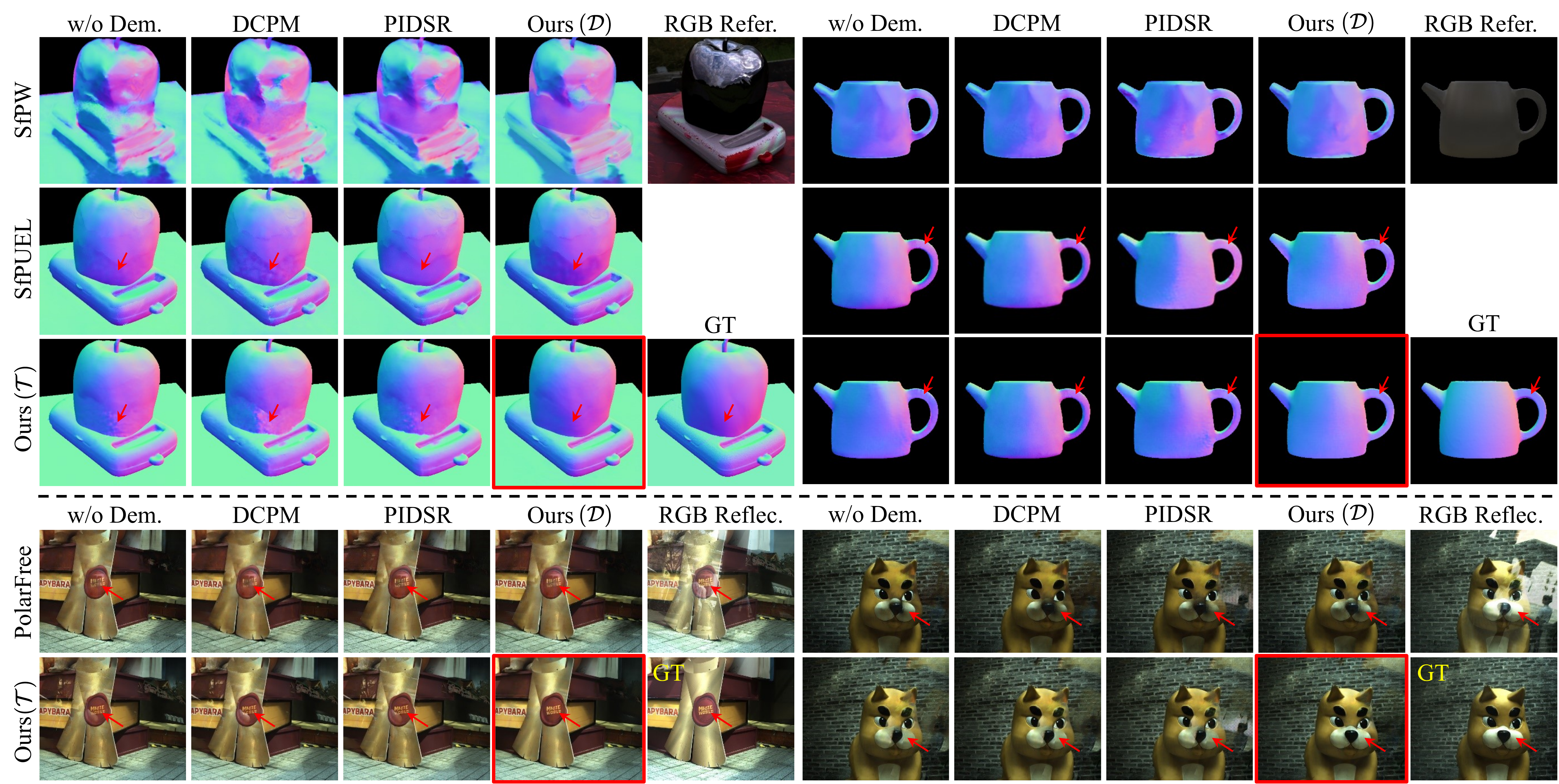}
    \captionsetup{type=figure,skip=4pt}
    \caption{Visual results in SfP (top) and DfP (bottom). Top: Our complete PolarAPP avoids artifacts in synthetic image (left) and ensure the normal consistency in real scene (right). Bottom: Our complete PolarAPP better removes reflection while retaining necessary details in reflective area. (\textbf{Zooming in for better comparison})}
\label{fig8}
\end{figure}}

\subsection{Task Comparison}

Our task comparisons include two settings. 
First, in simulation, we retain the imaging operator $\mathcal{A}$ to ensure compatibility with task labels from mosaic rearranged datasets. 
Second, in real-world experiments, we remove $\mathcal{A}$ to match the inference setup with practical deployment conditions.

The quantitative results are shown in Tab.~\ref{tab:2}, with the broader combination landscape in Fig.~\ref{fig7}. In Tab.~\ref{tab:2}, PIDSR+ denotes the two-stage baseline where the corresponding task network is retrained on outputs from a fixed PIDSR demosaicker, i.e., SfPUEL for SfP and PolarFree for DfP. PolarAPP achieves the best performance on all SfP metrics and the main DfP metrics, showing that the gain is not merely from using a stronger demosaicker or adapting the task network to PIDSR outputs. Fig.~\ref{fig7} further evaluates different demosaicker-task pairings as practical two-stage references. The results show that both modules of PolarAPP are individually effective: $\mathcal{D}$ provides task-effective polarization cues, and $\mathcal{T}$ remains robust to different upstream inputs. However, the best performance is achieved when both modules are trained together, confirming the benefit of task-aware coupling.

\begin{figure}[t]
\centering
\begin{minipage}[t]{0.48\textwidth}
    \null
    \centering
    \includegraphics[width=\linewidth]{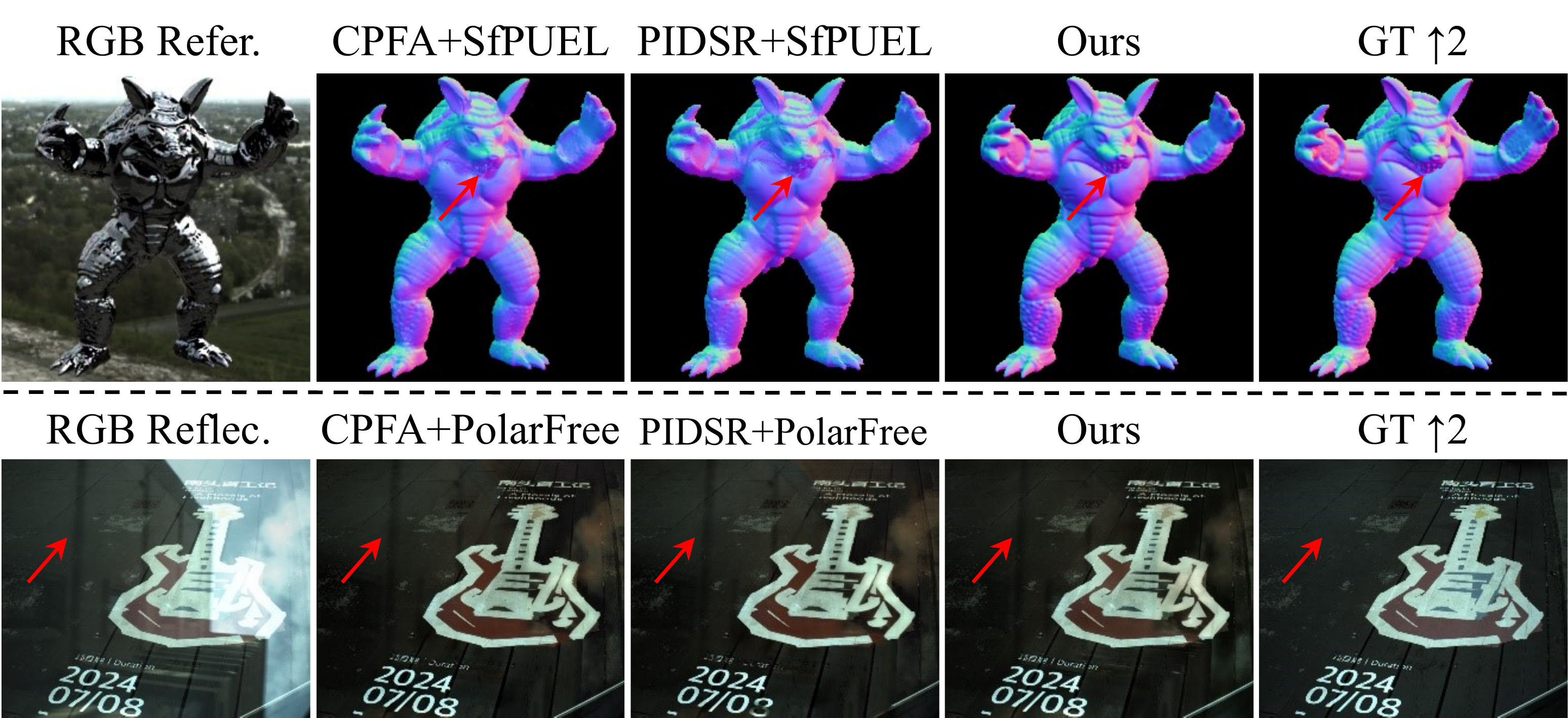}
    \captionsetup{type=figure,skip=-2pt}
    \captionof{figure}{Visual comparisons follow real DoFP imaging (without $\mathcal{A}$).}
    \label{fig9}
\end{minipage}\hfill
\begin{minipage}[t]{0.48\textwidth}
    \null
    \centering
    \includegraphics[width=\linewidth]{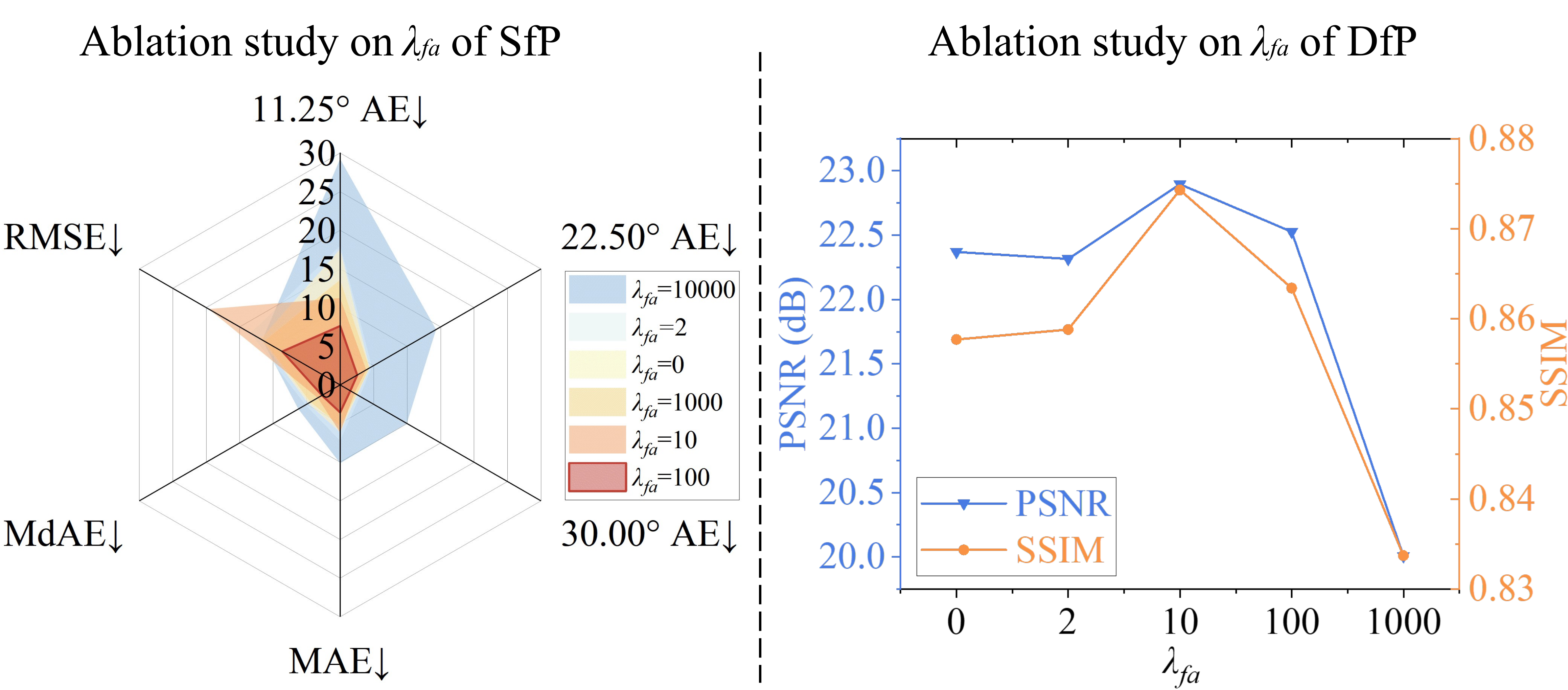}
    \captionsetup{type=figure,skip=0pt}
    \captionof{figure}{Ablation studies of $\lambda_{fa}$ on SfP (left) and DfP (right).}
    \label{fig10}
\end{minipage}
\end{figure}

Qualitative comparisons in Fig.~\ref{fig8} support these findings. In SfP scenes, alternative combinations produce distorted normals and inconsistent textures, while PolarAPP recovers correct orientations, sharper edges, and fewer shading artifacts. In DfP, PolarFree with other demosaickers suppresses reflections, and $\mathcal{T}$ paired with other demosaickers retains details, but only full PolarAPP removes reflections cleanly without losing fine structure. Real DoFP imaging results without $\mathcal{A}$ are shown in Fig.~\ref{fig9}. For direct task baselines such as SfPUEL and PolarFree, their predictions are bilinearly upsampled only for resolution alignment; in the PIDSR--SfPUEL and the PIDSR--PolarFree baseline, $\mathcal{A}$ is removed for consistency with the real setting. PolarAPP closely matches the GT on both SfP and DfP, whereas the other pipelines exhibit artifacts and texture degradation. Additional results are provided in the SM.

\subsection{Ablation study}

To evaluate the contribution of each component in PolarAPP, we conduct ablation studies on demosaicking and downstream tasks (SfP and DfP) under a unified setting that retains the imaging operator $\mathcal{A}$ during joint learning. Results are summarized in Tab.~\ref{tab:3}.

\noindent \textbf{$w/o$ $\mathcal{D}$ in training}. We remove the demosaicker and train the task network directly on CPFA-processed inputs, which matches the task-only pipeline used by SfPUEL in our comparisons. This variant leads to significant performance drop on both SfP and DfP, indicating that a task network alone cannot reliably compensate for the lack of explicit polarization demosaicking.

\noindent \textbf{Task-agnostic $\mathcal{D}$}. We further include a two-stage baseline where $\mathcal{D}$ is trained alone with $\mathcal{L}_d$ (including $\mathcal{L}_{eit}$) and then frozen while $\mathcal{T}$ is trained with $\mathcal{L}_t$ only. Compared with PolarAPP, this task-agnostic pipeline remains inferior, indicating that a fidelity-oriented $\mathcal{D}$ can be suboptimal for downstream utility even when $\mathcal{T}$ is trained on its fixed outputs. 
This supports the need for task-aware coupling between demosaicking and downstream prediction.

\noindent \textbf{$w/o$ $\mathcal{L}_{eit}$}. Removing the equivariant imaging transformation loss $\mathcal{L}_{eit}$ significantly reduces DoLP/AoP reconstruction quality and in turn harms downstream accuracy, confirming that $\mathcal{L}_{eit}$ provides effective self-supervision for stable polarization demosaicking under limited paired data.

\begin{table}[t]
\setlength{\belowcaptionskip}{3pt}
\caption{Quantitative results on ablation study.}
\label{tab:3}
\centering
\resizebox{\textwidth}{!}{
\begin{tabular}{c|ccccccccccc}
\hline
\multirow{2}{*}{Metrics} 
& \multicolumn{2}{c}{DoLP$\uparrow$} 
& AoP$\downarrow$
& \multicolumn{3}{c}{Normal Accuracy (\%)$\uparrow$} 
& \multicolumn{3}{c}{Normal Error ($^\circ$)$\downarrow$} 
& \multicolumn{2}{c}{De-reflection$\uparrow$} 
\\[2pt]

& PSNR  
& SSIM
& MAE  
& 11.25$^\circ$
& 22.5$^\circ$
& 30$^\circ$
& MAE
& MdAE
& RMSE
& PSNR
& SSIM
\\ \hline
\addlinespace[1pt]

$w/o$ $\mathcal{D}$ in training
& / & / & / 
& 78.14 & 91.11 & 95.97
& 6.01  & 5.61 & 13.40
& 22.26 & 0.851 
\\

Task-agnostic $\mathcal{D}$
& 38.13 & 0.927 & 5.502 
& 76.91 & 89.10 & 94.55
& 5.87  & 5.43 & 12.98
& 22.40 & 0.846 
\\

$w/o$ $\mathcal{L}_{eit}$ 
& 36.75 & 0.885 & 8.011 
& 86.12 & 92.43 & 95.75
& 5.73  & 4.96 & 12.57
& 22.46 & 0.869 
\\

Na\"ive Joint
& \textbf{38.43} & 0.924 & 5.658 
& 82.44 & 94.62 & 97.43
& 6.09  & 4.43 & 11.86
& 22.37 & 0.858 
\\

$w/o$ Refinement 
& 38.05 & 0.920 & 5.789 
& 89.01 & 93.55 & 96.33
& 4.68  & 4.50 & 10.13
& 22.59 & 0.869 
\\

Ours 
& 38.10 & \textbf{0.928} & \textbf{5.444} 
& \textbf{92.37} & \textbf{97.45} & \textbf{98.48}
& \textbf{3.57}  & \textbf{2.71} & \textbf{8.64}
& \textbf{22.90} & \textbf{0.871} 
\\ \hline

\end{tabular}}
\end{table}

\noindent \textbf{Na\"ive joint training ($w/o$ meta-learned alignment)}. We ablate the proposed meta-learning-based feature alignment by disabling it and performing standard end-to-end joint optimization of $\mathcal{D}$ and $\mathcal{T}$ using only $\mathcal{L}_d$+$\mathcal{L}_t$. This na\"ive baseline is consistently inferior to PolarAPP, demonstrating the value of goal-directed feature alignment.

\noindent \textbf{Effect of $\lambda_{fa}$}. We sweep the alignment weight $\lambda_{fa}$ in Fig.~\ref{fig10}. Performance peaks at $\lambda_{fa}=100$ for SfP and $\lambda_{fa}=10$ for DfP (for radar visualization, angular-accuracy thresholds are converted to angular-error thresholds so that lower is better). The different optima reflect task-dependent sensitivity: SfP benefits more from strong global geometric consistency, whereas DfP is more sensitive to fine appearance details, and overly strong alignment can interfere with task-specific objectives. In both cases, too small $\lambda_{fa}$ weakens coupling, while too large $\lambda_{fa}$ over-regularizes optimization.

\noindent \textbf{$w/o$ refinement}. Finally, the benefits of refinement are also reflected in improved metrics, indicating that this stage further unlocks the potential of the joint model. More discussion is provided in the SM, and we additionally evaluate a non-regression downstream task (material detection) in the SM.

\section{Conclusion}
In this work, we introduce PolarAPP, the first framework to jointly optimize polarization demosaicking and downstream tasks through task-specific joint learning. The proposed feature alignment mechanism, optimized through meta-learning, effectively bridges the objectives of the demosaicking and task networks and stabilizes their joint training. The demosaicking network is also guided by the equivalent imaging constraint, enabling effective regression learning without relying on low-quality datasets. Both innovations lead to superior performance in demosaicking and downstream tasks, including normal estimation and de-reflection. Experimental results demonstrate that PolarAPP outperforms existing methods, providing a flexible and robust solution for polarimetric imaging.

\section*{Acknowledgements}
This work was supported by the National Natural Science Foundation of China (grant number U2541205, 62271414), National Key R\&D Program of China (2024YFF0505603), ``Pioneer” and “Leading Goose” R\&D Program of Zhejiang (grant number 2024SDXHDX0006, 2024C03182), the 2023 International Sci-tech Cooperation Projects under the purview of the “Innovation Yongjiang 2035” Key R\&D Program (grant number 2024Z126), the National Natural Science Foundation of China (grant number 62105372) and Hunan Provincial Research and Development Project (grant number 2025QK3019).

\bibliographystyle{splncs04}
\bibliography{main}

\clearpage

% supplementary title
\appendix

% reset numbering
\setcounter{section}{0}
\setcounter{equation}{0}
\setcounter{figure}{0}
\setcounter{table}{0}

% redefine numbering format
\renewcommand{\thesection}{\Alph{section}}
\renewcommand{\theequation}{S\arabic{equation}}
\renewcommand{\thefigure}{S\arabic{figure}}
\renewcommand{\thetable}{S\arabic{table}}

\section{Theoretical Analysis of FT Update}

To better understand the learning behavior of the feature-alignment module, we analyze the optimization of ${FT_1, FT_2}$. During the outer update, their parameters $\theta_n$ ($n \in {1,2}$) are optimized using losses computed on the meta-testing set, while the resulting gradients must account for how the demosaicking network $\mathcal{D}$ and task network $\mathcal{T}$ responded to the meta-training set in the inner update. This results in a bi-level gradient consisting of two terms, one from each network.

\noindent \textbf{Gradient through Demosaicking Loss}. Let $\mathcal{L}_d^{mts}=\mathcal{L}_d(\mathcal{D}'(\mathcal{A}(\Imat_{mts}^{ref})),\Imat_{mts}^{ref})$ denote the demosaicking loss computed using the demosaicker $\mathcal{D}'$. The $FT$ parameters appear in this loss indirectly because: ($i$) $\theta_n$ affects $\mathcal{L}_{fa}$ in the inner update, ($ii$) $\mathcal{L}_{fa}$ affects the update of $\theta_{\mathcal{D}'}$ and ($iii$) $\theta_{\mathcal{D}'}$ affects the demosaicking loss on meta-testing. This produces the following gradient pathway:
\begin{equation}
    \textstyle \frac{\partial \mathcal{L}_d^{mts}}{\partial \theta_n}=\frac{\partial \mathcal{L}_d^{mts}}{\partial \theta_{\mathcal{D}'}} \cdot \frac{\partial \theta_{\mathcal{D}'}}{\partial \theta_n}.
    \label{eq:9}
\end{equation}
\noindent Using inner-update rule $\theta_{\mathcal{D}'}=\textstyle \theta_{\mathcal{D}} - \beta_{\mathcal{D}'} \frac{\partial \mathcal{L}_{fa}^{mtr}}{\partial \theta_{\mathcal{D}}}$, we obtain:
\begin{equation}
    \textstyle \frac{\partial \theta_{\mathcal{D}'}}{\partial \theta_n}=-\beta_{\mathcal{D}'} \frac{\partial^2 \mathcal{L}_{fa}^{mtr}}{\partial \theta_{n}\partial \theta_{\mathcal{D}}}.
    \label{eq:10}
\end{equation}
\noindent Substituting into Eq.~\eqref{eq:9} yields the demosaicking-related contribution:
\begin{equation}
    \textstyle -\beta_n \frac{\partial \mathcal{L}_d^{mts}}{\partial \theta_n}=\beta_n \beta_{\mathcal{D}'} \frac{\partial \mathcal{L}_d^{mts}}{\partial \theta_{\mathcal{D}'}} \cdot \frac{\partial^2 \mathcal{L}_{fa}^{mtr}}{\partial \theta_{n}\partial \theta_{\mathcal{D}}}.
    \label{eq:11}
\end{equation}

\noindent \textbf{Gradient through Task Loss}. Similarly, defining the task loss on the meta-testing set as $\mathcal{L}_t^{mts}=\mathcal{L}_t(\mathcal{T}'(\Imat_{mts}),\Ymat_{mts}^{gt})$. Since $\theta_n$ influences $\theta_{\mathcal{T}'}$ through the inner-update loss $\mathcal{L}_{fa}$, this gives:
\begin{equation}
    \textstyle \frac{\partial \mathcal{L}_t^{mts}}{\partial \theta_n}=\frac{\partial \mathcal{L}_t^{mts}}{\partial \theta_{\mathcal{T}'}} \cdot \frac{\partial \theta_{\mathcal{T}'}}{\partial \theta_n}.
    \label{eq:12}
\end{equation}
\noindent With inner-update rule $\theta_{\mathcal{T}'}=\textstyle \theta_{\mathcal{T}} - \beta_{\mathcal{T}'} \frac{\partial \mathcal{L}_{fa}^{mtr}}{\partial \theta_{\mathcal{T}}}$, we obtain:
\begin{equation}
    \textstyle \frac{\partial \theta_{\mathcal{T}'}}{\partial \theta_n}=-\beta_{\mathcal{T}'} \frac{\partial^2 \mathcal{L}_{fa}^{mtr}}{\partial \theta_{n}\partial \theta_{\mathcal{T}}},
    \label{eq:13}
\end{equation}
\noindent and thus yielding the task-loss contribution:
\begin{equation}
    \textstyle -\beta_n \frac{\partial \mathcal{L}_t^{mts}}{\partial \theta_n}=\beta_n \beta_{\mathcal{T}'} \frac{\partial \mathcal{L}_t^{mts}}{\partial \theta_{\mathcal{T}'}} \cdot \frac{\partial^2 \mathcal{L}_{fa}^{mtr}}{\partial \theta_{n}\partial \theta_{\mathcal{T}}}.
    \label{eq:14}
\end{equation}

Summing both pathways yields exactly the update rule in Eq. (3) (main text). This decomposition makes the learning roles clear: ($i$) the meta-testing losses $\mathcal{L}_d$ and $\mathcal{L}_t$ determine the update direction for the $FT$ modules, and ($ii$) the meta-training loss $\mathcal{L}_{fa}$ controls how strongly the $FT$ modules influence the adaptations of $\mathcal{D}$ and $\mathcal{T}$ through second-order gradients. Together, these interactions enable the $FT$ modules to learn alignment behaviors that improve joint optimization stability and downstream task performance.

\textit{Intuitively}, the $FT$ modules observe how changes in feature alignment, induced by $\mathcal{L}_{fa}$ on the meta-train set, modify the behavior of $\mathcal{D}$ and $\mathcal{T}$, and are then updated according to how well these adapted networks perform on the meta-test set. In this way, the $FT$ modules gradually learn alignment strategies that make the demosaicking and task networks more compatible within the joint training framework.

\section{Additional Implementation Details}

\begin{figure}[t]
\centering
    \includegraphics[width=\textwidth]{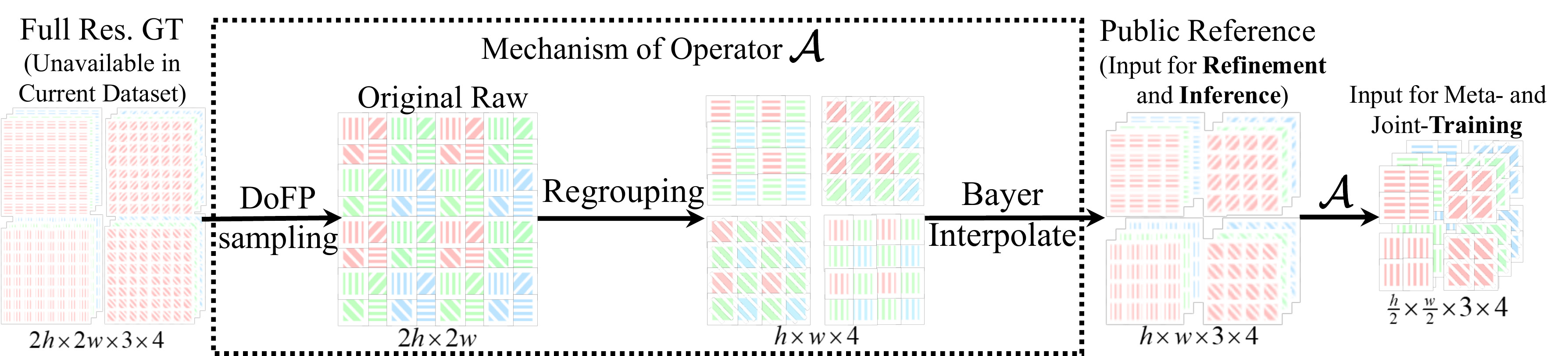}
    \caption{Resolution clarification of PolarAPP.}
\label{fig_res}
\end{figure}

\noindent \textbf{Resolution clarification of PolarAPP}. Fig.~\ref{fig_res} summarizes the resolution flow and data availability in PolarAPP. A DoFP CPFA sensor captures a raw mosaic at $2h\times2w$, whereas the corresponding full-resolution color polarization images ($2h\times2w\times3\times4$) are unknown in current public datasets. Instead, available “reference” polarization images are typically obtained by DoFP regrouping followed by Bayer demosaicking, and are provided at a unified spatial resolution of $h\times w$ ($h\times w\times3\times4$); the downstream task labels are also defined at $h\times w\times3$ of both SfP and DfP.

To construct paired supervision from task datasets, we introduce a deterministic imaging degradation operator $\mathcal{A}$ to synthesize DoFP-consistent low quality inputs from the reference images, thereby forming supervised pairs for training $\mathcal{D}$. Built on this paired construction, we further impose the EIT prior to enforce transformation-consistent demosaicking, which helps the model learn the underlying demosaicking process rather than overfitting dataset-specific statistics. In the refinement and inference stages, we remove $\mathcal{A}$ and feed the reference images directly; notably, $\mathcal{D}$ is frozen during refinement to bridge the train–test resolution gap. Inference ultimately produces full-resolution task outputs.

\noindent \textbf{Training time}. Different downstream tasks are trained separately due to their distinct objectives and training strategies. All experiments are conducted on 4 NVIDIA RTX 4090 GPUs with batch size 2 per GPU (global batch size 8) and input resolution $256\times256$. In each epoch, the meta-learning stage performs 200 fixed inner–outer cycles ($\sim5$ min), followed by joint learning and refinement over the full training set. One full training run requires roughly six/three days for SfP/DfP. 

\noindent \textbf{Efficiency}. We compares PolarAPP with other methods in terms of parameter count, FLoPs and runtime, and all the methods are evaluated on an NVIDIA RTX 4090 GPU. All demosaickers take inputs of size $512 \times 512$, while the task models operate on $1024 \times 1024$ inputs with a batch size of $1$ throughout. For runtime evaluation, we perform $100$ inference runs and report the average latency of the last $99$ iterations to exclude the warm-up overhead of the first run. As shown in Tab.~\ref{tab:4}, our demosaicker $\mathcal{D}$ achieves lightweight design—significantly reducing parameters, FLoPs, and latency compared to existing methods. In the SfP task, our task model $\mathcal{T}$ incurs only marginally higher computational cost than the underperforming SfPW baseline, yet uses fewer parameters. Thanks to the meta-learning-based training framework, PolarAPP maintains remarkably low computational complexity while delivering SOTA performance across demosaicking, SfP, and DfP tasks.

\begin{table}[t]
\centering
\caption{Model efficiency comparison. (Our $\mathcal{T}$ and PolarFree share the same network architecture in DfP, resulting in identical computational complexity).}
\label{tab:4}
\scalebox{.9}{
\begin{tabular}{ccccc}
\hline

\multicolumn{2}{c}{Metrics}      & Params (M)             & GFLoPs                  & Runtime (s)             \\ \hline
\addlinespace[1pt]

\multirow{3}{*}{Dem} & DCPM      & 573.75                 & 17936.02                & 0.5688                  \\
                     & PIDSR     & 7.45                   & 230.92                  & 0.1361                  \\
                     & Ours ($\mathcal{D}$)  & \textbf{4.40}                   & \textbf{117.83}                  & \textbf{0.0921}                  \\ \hline
\addlinespace[1pt]

\multirow{3}{*}{SfP} & SfPW      & 42.48                  & \textbf{187.33}                  & \textbf{0.1036}                  \\
                     & SfPUEL    & 138.33                 & 5303.67                 & 2.5441                  \\
                     & Ours ($\mathcal{T}$)  & \textbf{4.55}                   & 352.86                  & 0.2218                  \\ \hline
\addlinespace[1pt]

DfP                  & Ours ($\mathcal{T}$)  & 22.07                  & 460.21                  & 0.5077                  \\ \hline
\end{tabular}}
\end{table}

\section{Additional Ablations Study}

In Section 4.4, we analyze the contributions of individual modules in PolarAPP through controlled ablations. Although the refinement stage appears to have limited impact in simulation-based experiments, it provides clear benefits in real DoFP imaging. To highlight its practical value, we include an additional ablation study that directly follows the real imaging mechanism.

\begin{figure}[t]
\centering
    \includegraphics[width=\textwidth]{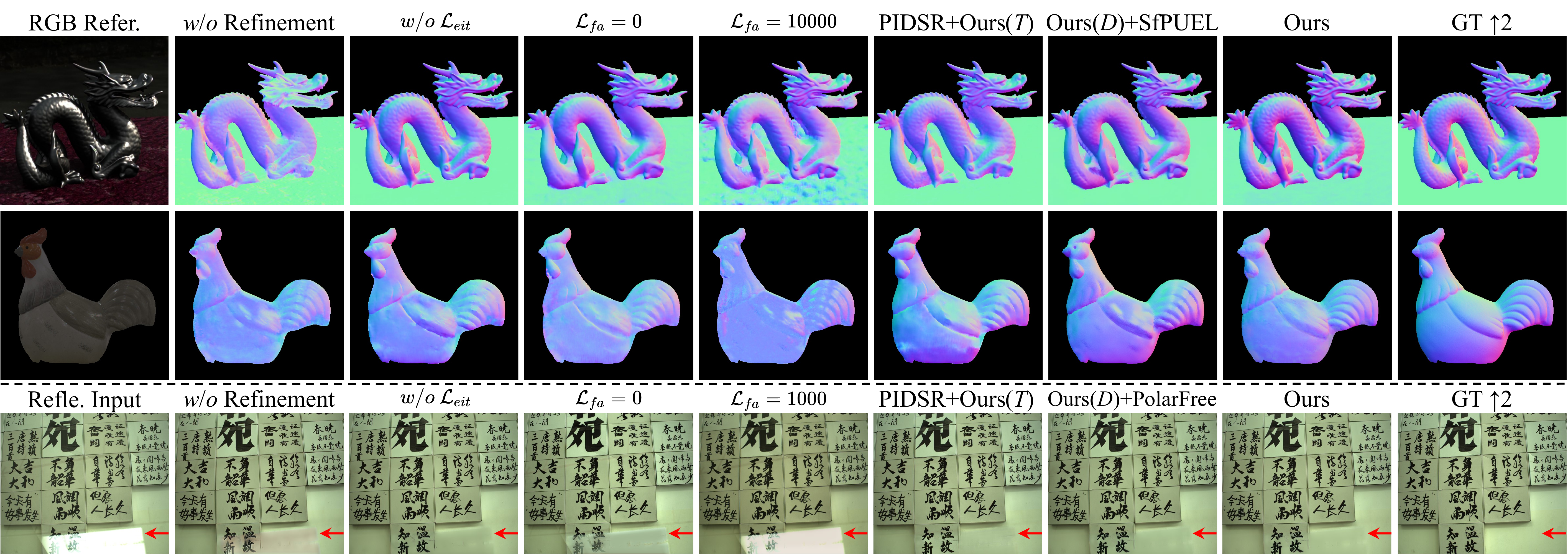}
    \caption{Visual results follow real DoFP imaging for ablation study.}
\label{fig10}
\end{figure}

\begin{figure}[h]
\centering
    \includegraphics[width=\columnwidth]{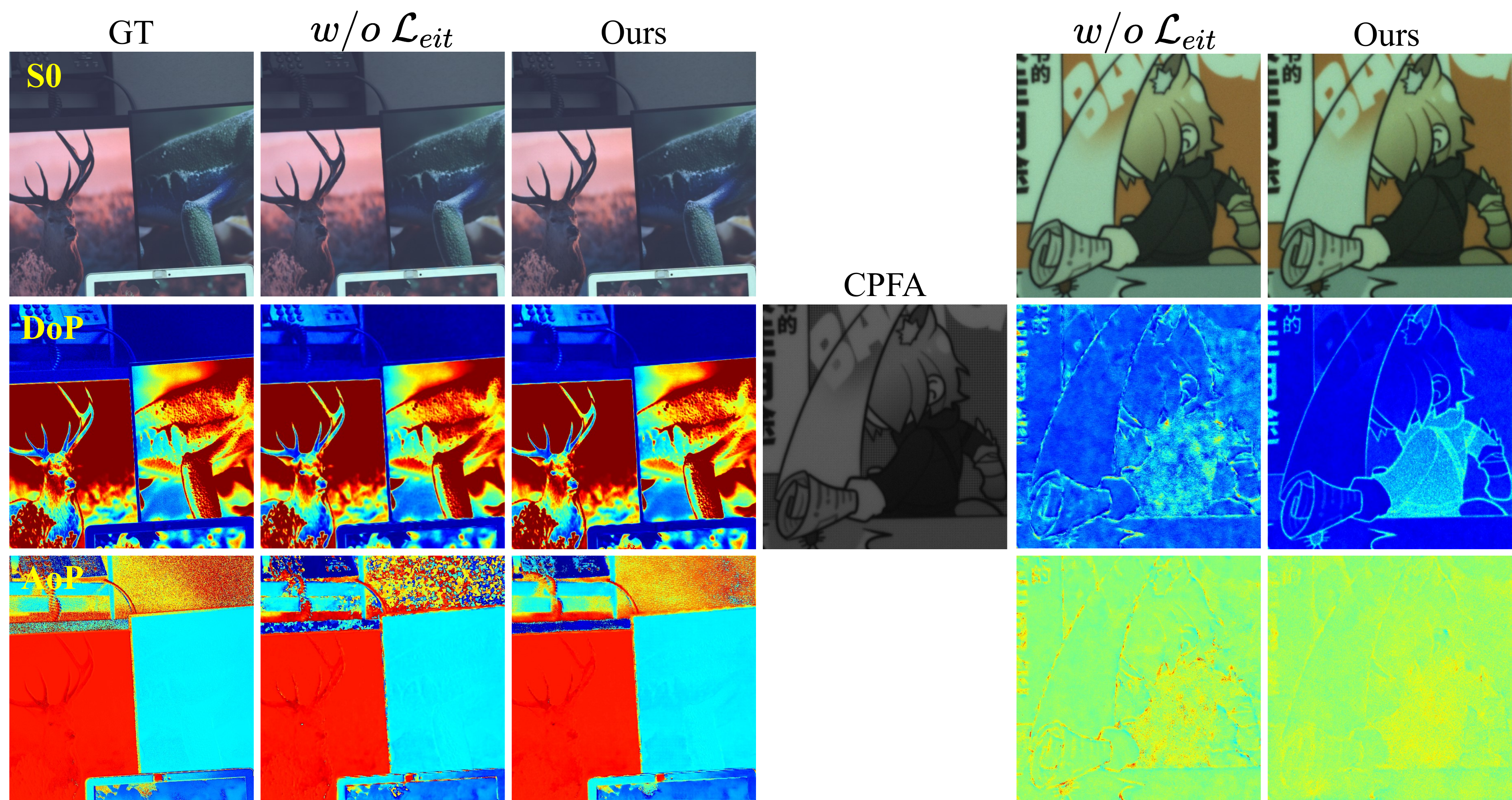}
    \caption{Visual results for ablation study of $\mathcal{L}_{eit}$.}
\label{fig11}
\end{figure}

Qualitative comparisons with and without refinement are shown in the second column of Fig.~\ref{fig10}. For SfP, both synthetic and real DoFP inputs reveal that removing refinement causes noticeable deviations from the upsampled ground truth. This reflects the strong sensitivity of normal estimation to resolution changes, since SfP depends on consistent angular cues across neighboring pixels. For DfP, the influence of refinement is reflected in the quality of reflection removal. Without refinement, PolarAPP suppresses specular reflections only partially, leaving visible color inconsistencies in reflective regions. These observations indicate that refinement plays an important role in restoring geometric accuracy for SfP and improving photometric consistency for DfP by compensating for resolution-related distortions and residual artifacts.

The remaining results in Fig.~\ref{fig10} illustrate the effect of removing other components under real DoFP imaging. Replacing the demosaicker with PIDSR or substituting the task networks with SfPUEL/PolarFree leads to inferior performance. Similar degradation appears when feature alignment is removed or excessively strengthened, or when the task network is trained independently. Although removing the EIT prior affects demosaicking more strongly than the downstream tasks, it still degrades normal estimation and de-reflection. As shown in Fig.~\ref{fig11}, retaining the EIT prior preserves more fine-scale details and provides cleaner polarization reconstructions.

\begin{figure}[h]
\centering
    \includegraphics[width=\textwidth]{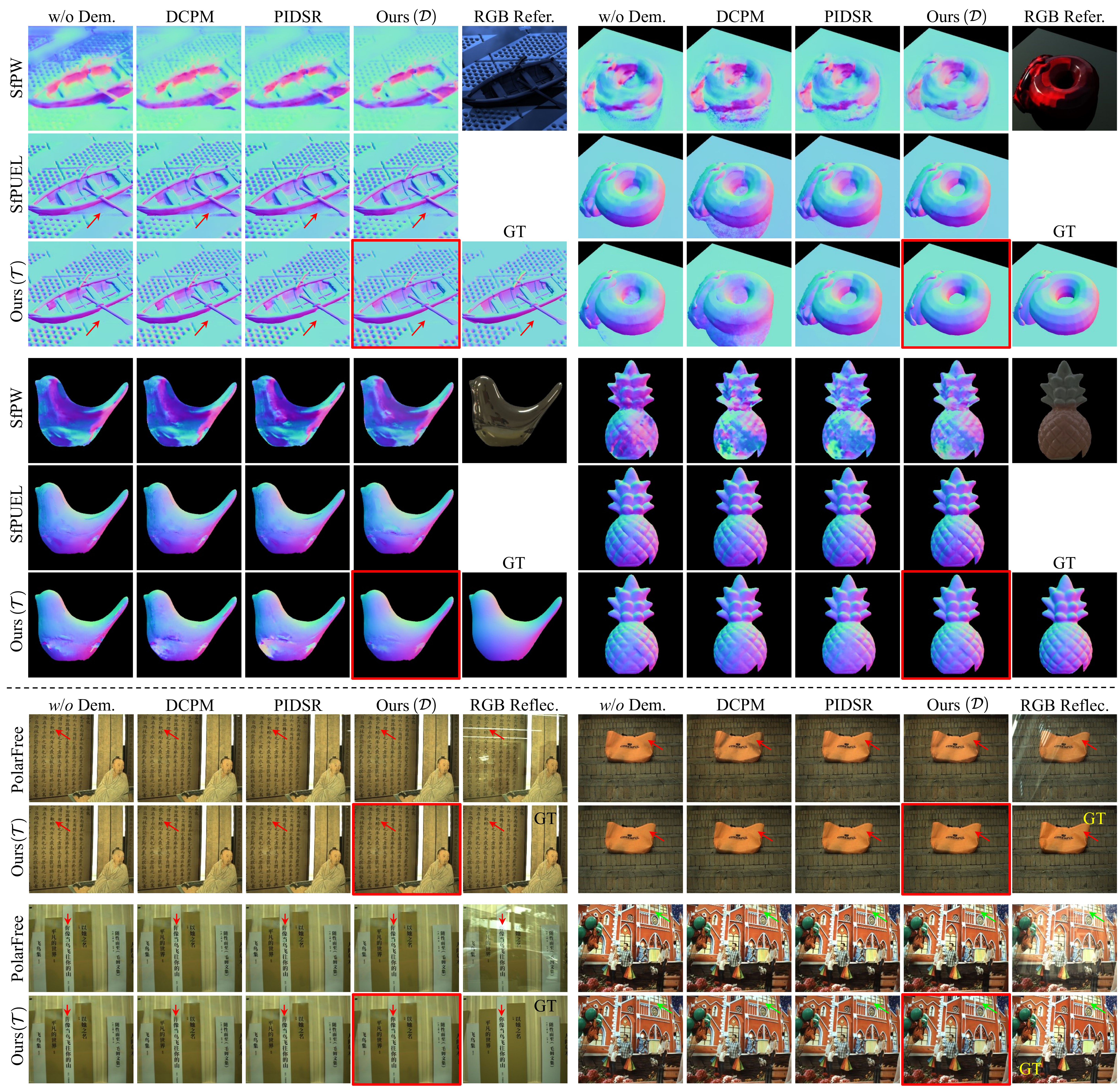}
    \caption{Additional visual results in SfP and DfP.}
\label{fig14}
\end{figure}

\begin{figure}[t]
\centering
    \includegraphics[width=\textwidth]{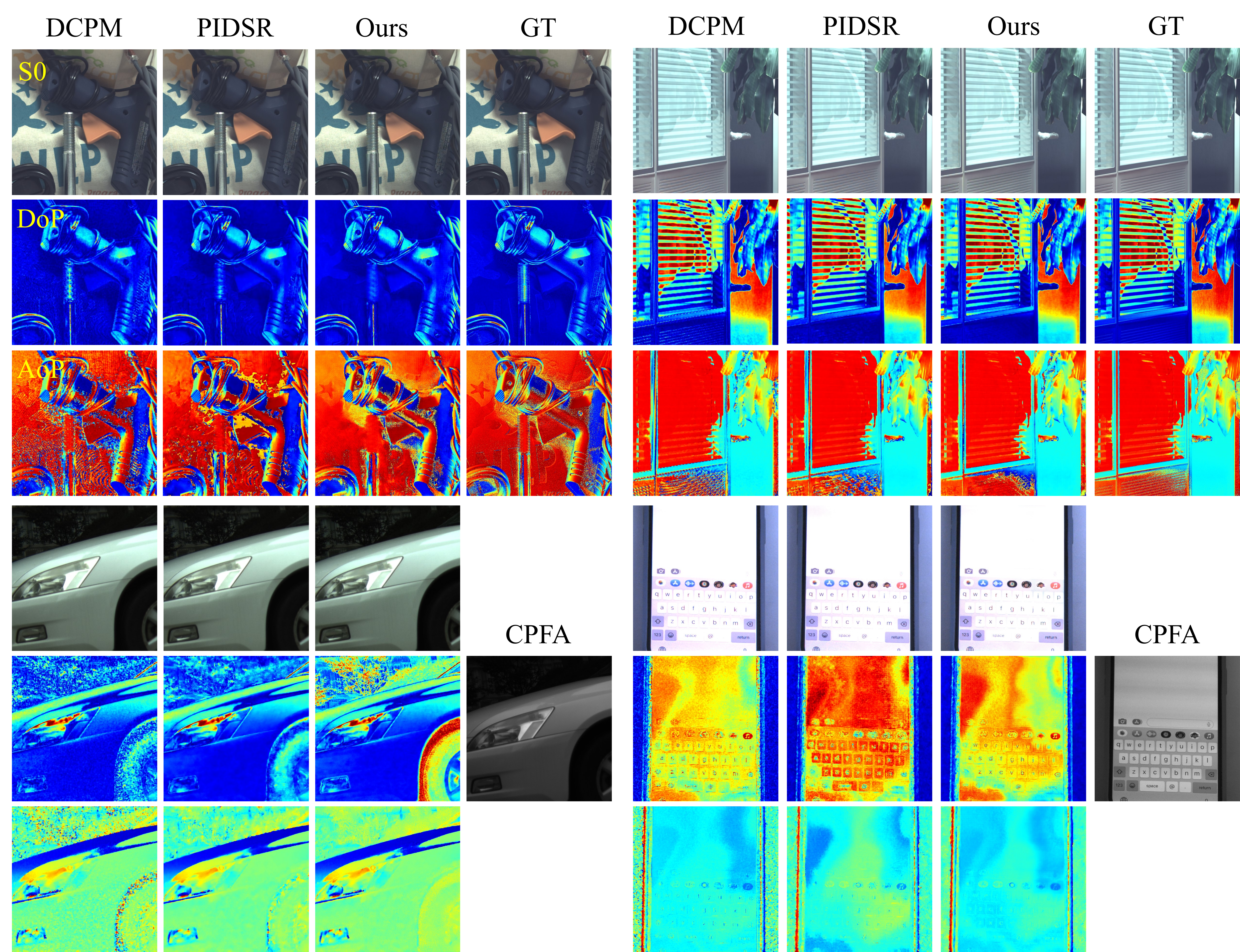}
    \caption{Additional visual results in demosaicking.}
\label{fig12}
\end{figure}
\begin{figure}[!h]
\centering
    \includegraphics[width=\textwidth]{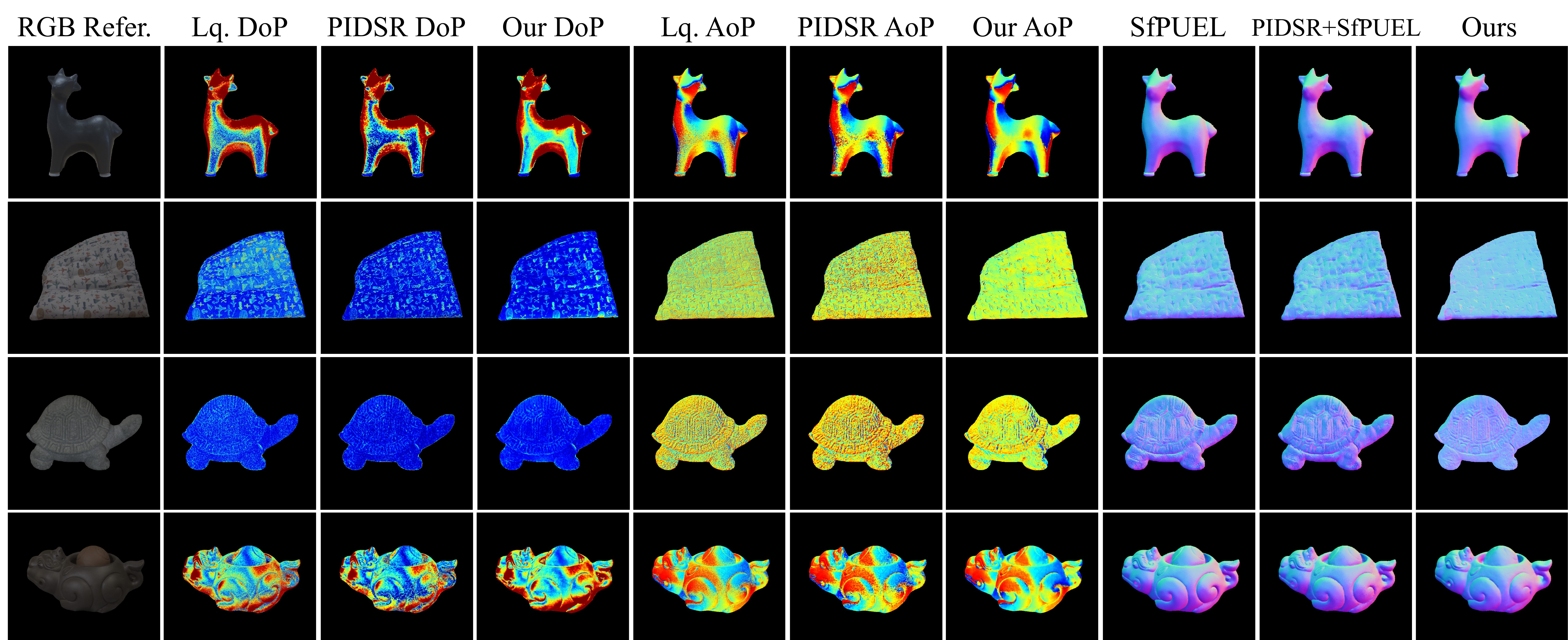}
    \caption{Visual results on Real-captured data without normal GT. Compared with DoLP$\&$AoP of low-quality ones and PIDSR, our clean DoLP$\&$AoP lead better normal estimation.}
\label{fig13}
\end{figure}

\section{Additional Visual Comparisons}

\begin{figure}[!t]
\centering
    \includegraphics[width=0.95\textwidth]{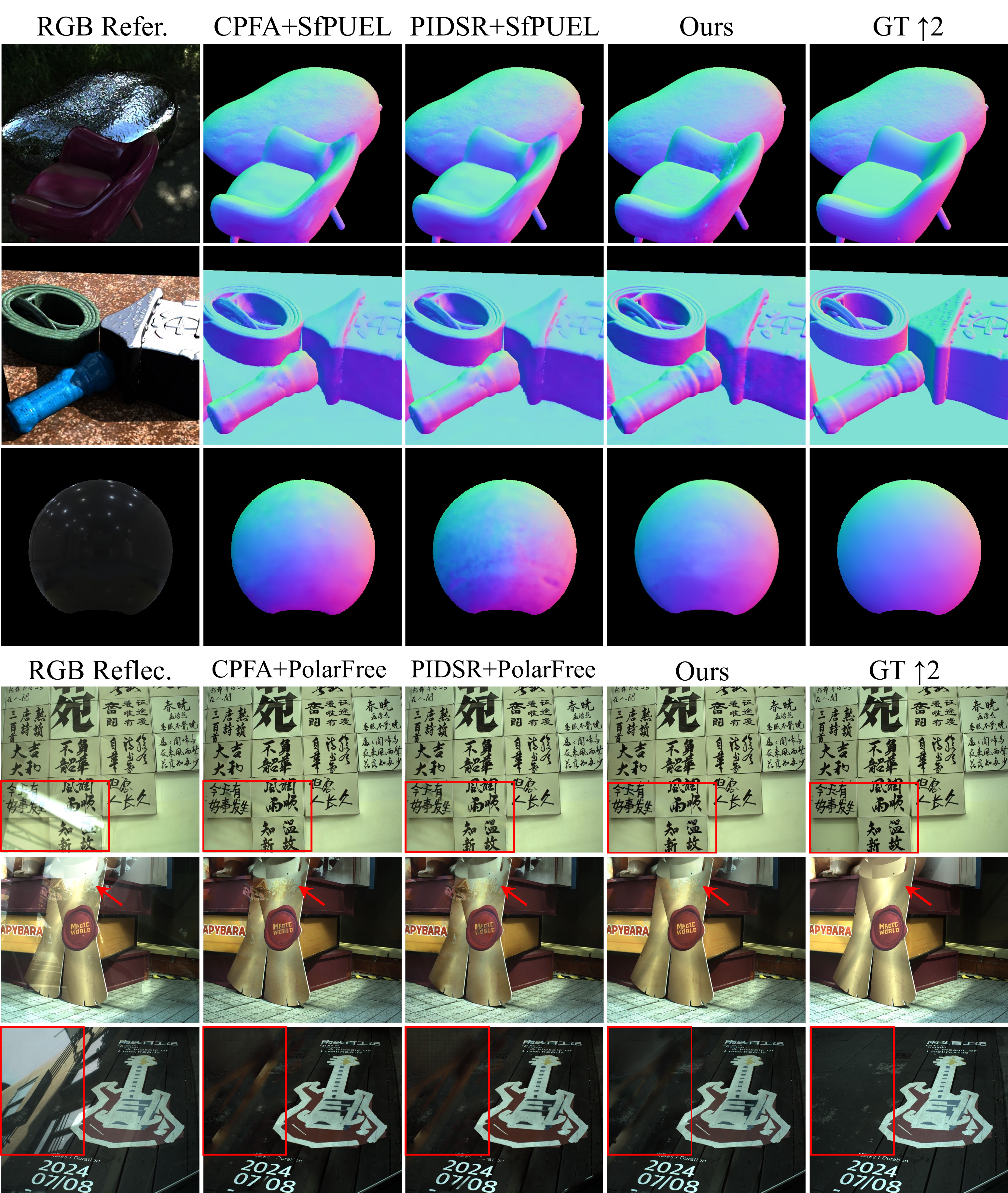}
    \caption{Additional visual results in SfP and DfP following the standard DoFP imaging.}
\label{fig15}
\end{figure}

This section presents additional visual results referenced in the main experiments, including comparisons for demosaicking, SfP, and DfP. Fig.~\ref{fig14} presents the SfP and DfP results when the imaging operator $\mathcal{A}$, is retained. Fig.~\ref{fig12} shows additional demosaicking results on the Qiu dataset and real-captured images. Fig.~\ref{fig13} compares PolarAPP with two other baseline methods on a real-captured dataset that lacks normal GT references. It demonstrates that our clean, low-noise DoLP and AoP are critical factors for generating higher-quality normal maps. Fig.~\ref{fig15} displays the corresponding results under the standard DoFP imaging pipeline without using $\mathcal{A}$. 

\begin{figure}[t]
\centering
    \includegraphics[width=\textwidth]{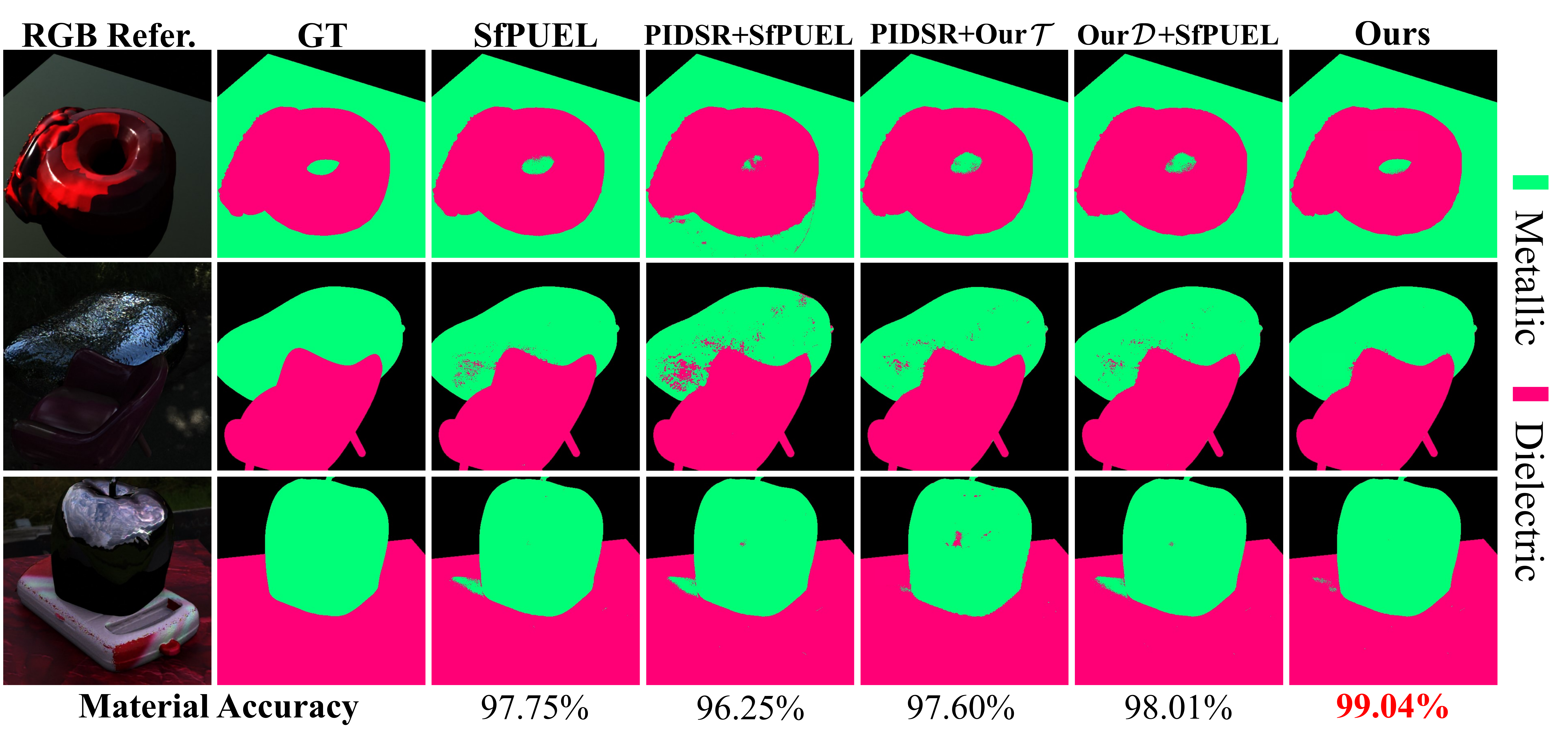}
    \caption{Visual comparisons on material detection. The detection accuracy of each pipeline is listed below the results, and PolarAPP still achieves the highest.}
\label{fig_seg}
\end{figure}

\section{Task generalization}
Since SfP and DfP are both regression-based tasks, we further evaluate the generality of PolarAPP on a \emph{non-regression} downstream task, namely material detection, which classifies dielectric and metallic materials from polarization cues. We use the dataset released with SfPUEL, where 19,800 out of 20,000 image groups are used for training and the remaining 200 for testing, and adopt the same task network $\mathcal{T}$ as in SfP. For supervision, we use the standard cross-entropy loss for two-class prediction, and set $\lambda_{fa}=100$.

For fair comparison, we evaluate five representative pipelines: ($i$) \textit{SfPUEL}. The task-only baseline that inputs CPFA directly. ($ii$) \textit{PIDSR+SfPUEL}. Task-agnostic demosaicking followed by a separately trained task network. ($iii$) \textit{$\mathcal{D}$ +SfPUEL}. Our demosaicker combined with the external task network. ($iv$) \textit{PIDSR+$\mathcal{T}$}. External demosaicker combined with our task network. ($v$) Our complete PolarAPP. Notably, the material labels are discrete binary maps, so their $2\times$ supervision can be obtained by lossless upsampling without introducing interpolation ambiguity, allowing direct quantitative evaluation at the refined resolution. As shown in Fig.~\ref{fig_seg}, the full PolarAPP achieves the highest material detection accuracy among all compared settings, indicating that the proposed joint structure generalizes beyond regression tasks and remains beneficial for discrete polarimetric perception.

\section{Discussion}

PolarAPP provides a unified framework for polarization demosaicking and downstream tasks, but several limitations remain. First, although PolarAPP achieves consistent gains across all tasks, its ultimate performance is influenced by the strength of the underlying demosaicking and task architectures. PolarAPP improves their interaction through feature alignment and meta-learning, but further advances in backbone design or physically grounded architectures may unlock additional improvements. Second, publicly available polarimetric datasets are limited in quality and diversity; many rely on regrouped mosaics or provide incomplete ground truth, which restricts the attainable accuracy of both reconstruction and downstream predictions. Third, this work trains task-specific $\mathcal{D}$--$\mathcal{T}$ pairs for SfP and DfP because their task losses and preferred polarization cues differ. A shared demosaicker with multiple task heads can be incorporated into the same framework by aggregating task losses, but balancing potentially conflicting task gradients is left for future work. Fourth, our current study isolates mosaic-induced degradation and does not explicitly model shot noise or readout noise in CPFA measurements. Extending PolarAPP to noisy raw observations is an important future direction.

Looking forward, the increasing availability of compact DoFP sensors and the growing interest in polarization-based vision suggest opportunities for more comprehensive datasets and physics-aware architectures. We expect future research to explore a broader spectrum of polarimetric applications, including underwater imaging, dehazing, transparent and specular scene perception, low-visibility target sensing, remote sensing, biomedical imaging, and industrial inspection. These directions further highlight the importance of polarization cues for robust visual understanding in challenging real-world environments, as well as the need for unified and task-optimized polarimetric frameworks.

\end{document}